\documentclass[11pt]{article}

\usepackage[final]{acl}

\usepackage{times}
\usepackage{latexsym}
\usepackage[T1]{fontenc}
\usepackage[utf8]{inputenc}
\usepackage{microtype}
\usepackage{inconsolata}
\usepackage{graphicx}
\usepackage{amsmath}
\usepackage{amssymb}
\usepackage{url}
\usepackage{array}
\usepackage{booktabs}
\usepackage{multirow}
\usepackage[multiple]{footmisc}

\usepackage{booktabs}
\usepackage{tabularx}

\title{When Modality Gap Reduction Fails: Prediction-Level Hubness in CLIP}

\author{
\textbf{Shota Sato}, \textbf{Hajime Kiyama}, \textbf{Tosho Hirasawa}, \textbf{Mamoru Komachi} \\[0.5em]
Hitotsubashi University \\
\texttt{\{shota,hajime,tosho,komachi\}@scl.sds.hit-u.ac.jp}
}

\begin{document}
\maketitle

\begin{abstract}
Reducing the modality gap between image and text representations in CLIP is widely expected to improve cross-modal alignment and downstream performance.
However, a smaller average image--text gap does not necessarily lead to consistent accuracy gains.
We analyze this mismatch from the perspective of the decision structure in zero-shot classification, i.e. selecting the most similar class-text prototype for an input image.
Zero-shot accuracy depends not only on average image--text alignment, but also on class-wise decision margins.
Using Linear correction as an analytically tractable case, we show that modality gap correction can alter the relative decision structure among classes and cause predictions to concentrate on a small subset of classes.
We refer to this output-space failure mode as \textit{prediction-level hubness}.
Furthermore, experiments across multiple datasets show that accuracy degradation under gap correction is consistently associated with increased prediction concentration, both for Linear correction and for learning-based correction methods.
This provides a systematic explanation of why modality gap reduction does not consistently improve CLIP zero-shot accuracy from the perspective of downstream decision structure.
Our results suggest that gap correction should be evaluated not only by average alignment, but also by its impact on downstream prediction structure.
\end{abstract}

\section{Introduction}
\label{sec:introduction}

\begin{figure}[t]
\centering
\includegraphics[keepaspectratio, scale=0.13]{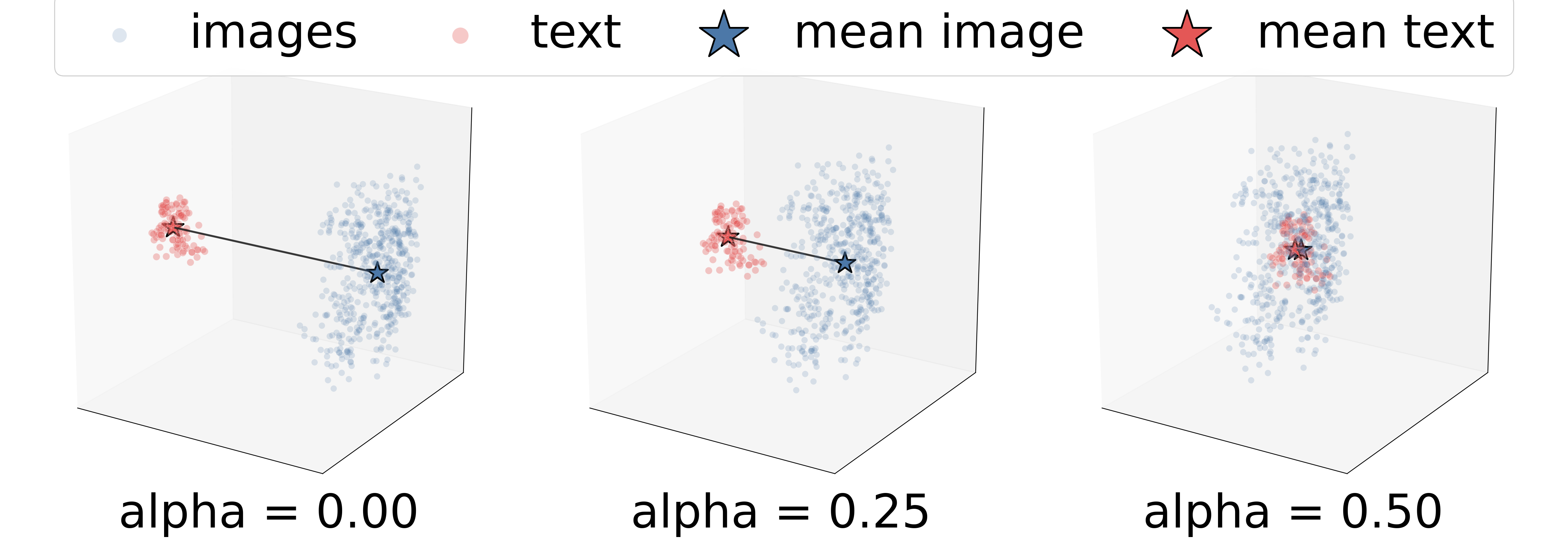}
\caption{PCA visualization of image and text embeddings under modality gap reduction. As the correction strength $\alpha$ increases, the mean image and text embeddings move closer, reducing the average modality gap.}
\label{fig:pca_plot}
\end{figure}

Contrastive Language--Image Pre-training (CLIP)~\cite{CLIP} learns a shared image--text embedding space and enables strong zero-shot transfer.
However, CLIP is known to exhibit a \textit{modality gap}, where image and text representations form modality-specific clusters~\cite{modality-gap}.
This has motivated methods for reducing the gap, including post-hoc geometric correction based on an estimated gap vector~\cite{modality-gap}, additional training such as CLIPRefine~\cite{CLIPRefine}, and alignment-oriented pre-training mechanisms such as AlignCLIP~\cite{AlignCLIP}.
Figure~\ref{fig:pca_plot} illustrates this geometric view of Linear correction, where increasing the correction strength brings the mean image and text embeddings closer.
These methods are often motivated by the expectation that improving average image--text alignment, operationalized as reducing the squared norm of the gap vector defined in Equation~\ref{eq:gap_vector}, improves downstream performance.

\begin{figure*}[t]
    \centering
    \includegraphics[keepaspectratio, scale=0.35]{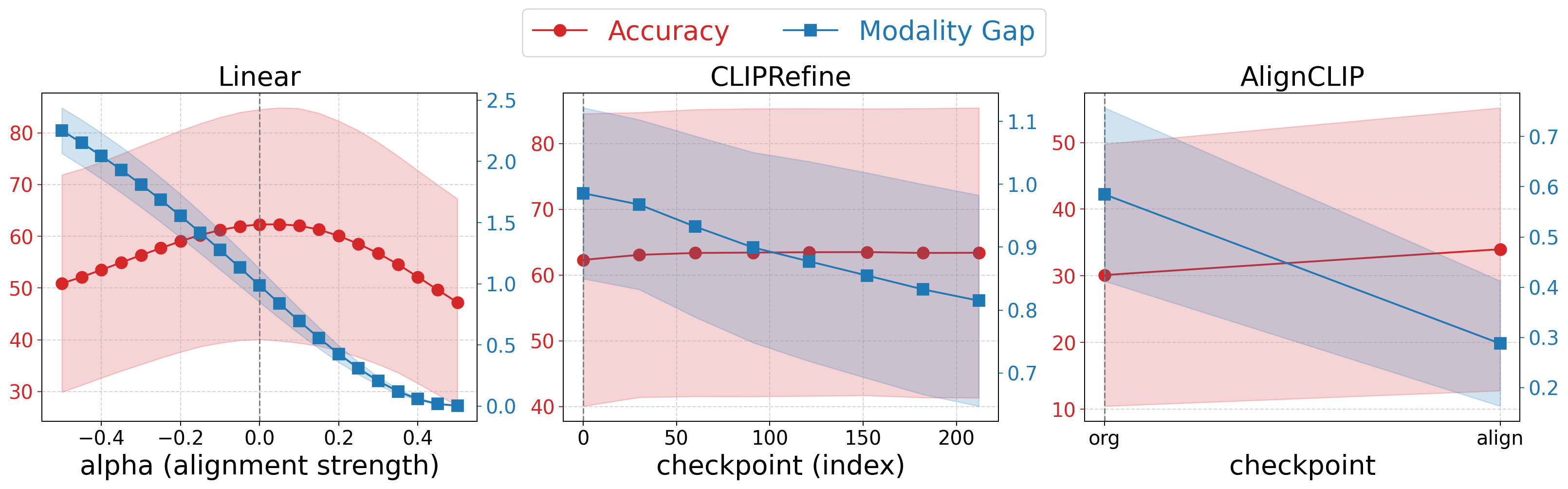}
    \caption{
    Accuracy and modality gap under modality gap reduction.
    Solid lines indicate the mean over 10 zero-shot image classification datasets, and shaded regions indicate the standard deviation.
    Under Linear correction, the modality gap decreases monotonically with correction strength, yet accuracy eventually deteriorates.
    In contrast, learning-based correction methods reduce the modality gap while maintaining or improving accuracy.
    }
    \label{fig:rawdual}
\end{figure*}

However, reducing the average modality gap does not necessarily yield consistent downstream improvements~\cite{modality-gap,Jiang_2023_CVPR,closinggap}.
As shown in Figure~\ref{fig:rawdual}, Linear correction monotonically reduces the modality gap across 10 zero-shot image classification datasets, yet accuracy eventually deteriorates.
This suggests that reducing the mean image--text displacement alone is insufficient to explain downstream performance changes, because zero-shot accuracy also depends on class-wise decision margins.
When gap reduction unevenly changes these margins, predictions may concentrate on a small subset of classes.

We analyze this effect using Linear correction as an analytically tractable case.
We show that the gap vector introduces unequal score shifts across classes, which can systematically favor certain class prototypes and lead corrected predictions to concentrate on them.
We call this output-space failure mode \textit{prediction-level hubness}: a \textit{prediction hub} is a class label that receives a disproportionate share of the final predicted labels after gap correction---distinct from classical nearest-neighbour hubness in embedding space~\citep{hubs-in-space}.

To test this mechanism, we examine whether the predicted class distribution changes in the way suggested by the above analysis.
We analyze predicted-class distributions across datasets and correction methods, and quantify how strongly predictions concentrate on a small subset of classes.
Consistent with the proposed mechanism, excessive Linear correction increases prediction concentration and degrades accuracy.
For learning-based corrections, accuracy degradation is also associated with increased prediction concentration, suggesting that prediction-level hubness is a useful diagnostic for failed modality gap reduction (correlational rather than mechanistic; Section~\ref{sec:mechanism}).

This work explains why modality gap reduction does not consistently improve CLIP zero-shot accuracy from the perspective of downstream decision structure and predicted-class distributions.
Our findings suggest that gap correction should be evaluated not only by average image--text alignment, but also by its impact on downstream prediction structure (practical guidance in Section~\ref{subsec:reporting}).

Our contributions are as follows:
\begin{enumerate}
    \item  We show that, in zero-shot classification, reducing the average modality gap does not reliably improve accuracy, even when the gap decreases monotonically.
    \item We analyze this mismatch through zero-shot prediction structure and introduce \textit{prediction-level hubness} as an output-space failure mode, using Linear correction as an analytically tractable case.
    \item We link prediction concentration to accuracy degradation across Linear and learning-based corrections, and show that gap-induced class-wise bias helps explain hub formation under Linear correction.
\end{enumerate}
\section{Background and Related Work}
\label{sec:related_work}

\paragraph{CLIP and the Modality Gap}

Contrastive Language--Image Pre-training (CLIP)~\cite{CLIP} learns a shared image--text embedding space using contrastive learning.
It enables strong zero-shot image classification by comparing image embeddings with class-text prototypes.
CLIP has also been widely used as a visual backbone in vision--language models~\cite{BLIP-2,LLaVa,LLaVa-1.5,MLLM_survey_2024}, and its contrastive learning framework has been extended to other modalities, such as audio--language learning~\cite{CLAP}.

Despite its joint embedding framework, CLIP is known to exhibit a \textit{modality gap}, where image and text embeddings form modality-specific clusters~\cite{modality-gap}.
This phenomenon has been analyzed from several perspectives, including representation geometry, contrastive learning dynamics, and modality-specific structure~\cite{modality-gap,towards-understanding,Cross-Uniformity,two-effect,pmlr-v280-yaras25a}.

\paragraph{Modality Gap Reduction}

Many methods have been proposed to reduce the modality gap in CLIP-like embedding spaces.
Post-hoc methods reduce the gap by applying geometric or statistical transformations after pre-training~\cite{modality-gap,procrustes,I0T,Fill-the-gap}.
Other methods refine the embedding space through additional training, regularization, or adversarial objectives~\cite{DIAS,Cross-Uniformity,MLP+GRL,CLIPRefine}.
Another line of work incorporates alignment-oriented mechanisms during pre-training or model design~\cite{D-Bridge,AlignCLIP,pmlr-v280-yaras25a}.

These studies generally aim to improve average image--text alignment by reducing modality separation.
However, prior work has also suggested that reducing the modality gap does not always lead to better downstream performance~\cite{modality-gap,Jiang_2023_CVPR,closinggap}.
Building on these observations, we study this gap--performance mismatch more broadly across multiple datasets and correction methods.

\paragraph{Hubness in Representation Spaces}

Hubness was originally studied as a high-dimensional nearest-neighbor phenomenon, where a small number of points appear disproportionately often in other points' nearest-neighbor lists~\citep{hubs-in-space,local-global-scaling,hubness-reduction-survey}.
This issue has also been discussed in zero-shot learning and cross-space mapping, where a few semantic prototypes or labels can become frequent nearest-neighbor targets~\citep{zero-shot-hub,hubness-pollution}.
In cross-lingual and cross-modal retrieval, related work has examined how such hubs affect retrieval quality and how local scaling or neighborhood-based methods can mitigate them~\citep{CSLS,balance-act,NeighborRetr}.
Recent studies have further analyzed hubness and vulnerability in CLIP-like representation spaces~\citep{deguchi-etal-2026-one}.

These studies primarily characterize hubness through nearest-neighbor relations in embedding or retrieval spaces.
In this work, we use the related but distinct notion of concentration at the level of final zero-shot predictions, focusing on how modality gap correction changes the predicted-class distribution.
\section{Problem Setup}
\label{sec:problem_setup}

This section introduces the problem addressed in this work: reducing the modality gap does not necessarily improve downstream performance.
In particular, we examine whether making image and text representations closer on average leads to better zero-shot image classification accuracy.
We use zero-shot classification as an evaluation setting because it directly exposes both image--text alignment and class-level prediction behavior.

\subsection{Average Modality Gap}

Let \(x_i \in \mathbb{R}^d\) and \(t_j \in \mathbb{R}^d\) denote the \(i\)-th image embedding and the \(j\)-th text embedding, respectively.
Following the gap-vector estimate used in Linear correction~\cite{modality-gap}, we define the gap vector as
\begin{equation}
g
=
\frac{1}{N}\sum_{i=1}^{N} x_i
-
\frac{1}{N}\sum_{j=1}^{N} t_j ,
\label{eq:gap_vector}
\end{equation}
where \(N\) is the number of embeddings averaged in each modality.
We define the average modality gap as the squared Euclidean distance between the mean image embedding and the mean text embedding:
\begin{equation}
\mathrm{MG}
=
\left\|
\frac{1}{N}\sum_{i=1}^{N} x_i
-
\frac{1}{N}\sum_{j=1}^{N} t_j
\right\|_2^2
=
\|g\|_2^2 .
\label{eq:modality_gap}
\end{equation}
This metric measures the average displacement between image and text representations in the shared embedding space.
A smaller value indicates that the two modalities are closer on average, and such a reduction is often expected to reflect better image--text alignment and downstream performance.
However, this expectation does not necessarily hold in downstream evaluation.
A smaller average modality gap does not guarantee that the task-relevant decision structure is improved.
In zero-shot image classification, predictions depend on relative scores between an image and all class-text prototypes.
Thus, even if the average image--text gap decreases, the score ordering among classes can change in a way that hurts accuracy.

\subsection{Gap Reduction and Accuracy}

We evaluate this issue using three representative correction profiles.
Linear correction is a post-hoc geometric correction based on an estimated gap vector~\cite{modality-gap}.
CLIPRefine reduces the gap through additional training after pre-training~\cite{CLIPRefine}.
AlignCLIP incorporates alignment-oriented mechanisms during pre-training~\cite{AlignCLIP}.\footnote{Detailed implementation settings, including prompt templates, correction strengths, checkpoints, and backbone differences, are provided in Appendix~\ref{app:additional_experimental_details}.}

Figure~\ref{fig:rawdual} summarizes the relationship between modality gap and accuracy for these profiles.
For Linear correction, the modality gap decreases monotonically as the correction strength increases, but accuracy improves only up to a point and then deteriorates.
In contrast, learning-based corrections reduce the modality gap while maintaining or improving accuracy.
These results show that modality gap reduction and accuracy improvement are not equivalent.

This mismatch motivates the following analysis.
If reducing the average modality gap is not sufficient to explain performance changes, we need to examine how gap correction changes prediction behavior.
We next provide a margin-based analysis of this mismatch and introduce \textit{prediction-level hubness} as a prediction-level failure mode.

\section{Mechanism: Gap-Induced Class-wise Bias and Prediction-Level Hubness}
\label{sec:mechanism}

We explain why reducing the average modality gap can fail to improve zero-shot accuracy.
The key point is that the modality gap is an average geometric quantity, whereas classification accuracy depends on class-wise decision margins.
We then show that Linear correction induces an explicit class-wise score bias, and empirically verify that this bias predicts which classes become prediction hubs.

Our analyses carry different evidential weight: Section~\ref{subsec:margin_view} is \textbf{general} (any correction that shifts zero-shot scores), whereas Section~\ref{subsec:linear_bias_hubs} is \textbf{mechanistic but Linear-specific}.

\subsection{Accuracy Depends on Class-wise Margins}
\label{subsec:margin_view}

Let \(s_{i,c}^{(k)}\) denote the zero-shot score between image \(i\) and class prototype \(c\) after correction step \(k\).
We define the score shift induced by correction as
\[
\delta_{i,c}^{(k)}
=
s_{i,c}^{(k)} - s_{i,c}^{(0)} .
\]
For the ground-truth class \(y_i\) and a competing class \(c\), the decision margin at correction step \(k\) is
\[
\Delta_{i,c}^{(k)}
=
s_{i,y_i}^{(k)} - s_{i,c}^{(k)} .
\]
Substituting the score shifts gives
\[
\begin{aligned}
\Delta_{i,c}^{(k)}
&=
\left(s_{i,y_i}^{(0)}+\delta_{i,y_i}^{(k)}\right)
-
\left(s_{i,c}^{(0)}+\delta_{i,c}^{(k)}\right) \\
&=
\Delta_{i,c}^{(0)}
+
\delta_{i,y_i}^{(k)}
-
\delta_{i,c}^{(k)} .
\end{aligned}
\]
Thus, a previously correct prediction can change to class \(c\) when
\[
\delta_{i,c}^{(k)}
-
\delta_{i,y_i}^{(k)}
>
\Delta_{i,c}^{(0)} .
\]
This condition depends on the relative score shifts of the ground-truth and competing classes.
It is therefore possible for correction to reduce the average modality gap while still decreasing accuracy, if competing classes receive larger score gains than the ground-truth class.


\begin{table}[t]
\centering
\small
\setlength{\tabcolsep}{4pt}
\begin{tabular}{lccc}
\toprule
Dataset
& \(\rho_m\)
& \(\rho_{\Delta m}\)
& \(\rho_e\) \\
\midrule
Caltech101    & 0.821 & 0.824 & 0.792 \\
CIFAR-10      & 0.891 & 0.903 & 0.819 \\
CIFAR-100     & 0.869 & 0.863 & 0.833 \\
DTD           & 0.586 & 0.918 & 0.770 \\
EuroSAT       & 0.612 & 0.976 & 0.796 \\
FGVC-Aircraft & 0.570 & 0.531 & 0.585 \\
Flowers102    & 0.747 & 0.616 & 0.802 \\
Food-101      & 0.747 & 0.817 & 0.842 \\
ImageNet-1K      & 0.835 & 0.812 & 0.748 \\
Oxford Pets   & 0.749 & 0.783 & 0.793 \\
\midrule
Mean          & 0.743 & 0.804 & 0.778 \\
\bottomrule
\end{tabular}
\caption{
Class-wise gap-induced bias predicts prediction hubs and harmful transition destinations under Linear over-correction.
We report Spearman correlations between \(b_c=-\langle g,t_c\rangle\) and three class-level quantities:
\(\rho_m=\rho(b_c,m_c^{(\alpha)})\),
\(\rho_{\Delta m}=\rho(b_c,\Delta m_c^{(\alpha)})\), and
\(\rho_e=\rho(b_c,e_c^{(\alpha)})\).
All correlations are computed at \(\alpha=0.5\).
}
\label{tab:linear_bias_hub_corr}
\end{table}

\subsection{Gap-Induced Bias Creates Prediction-Level Hubness}
\label{subsec:linear_bias_hubs}

For Linear correction, the general correction setting \(k\) is instantiated by the correction strength \(\alpha\).
Linear correction shifts image and text embeddings in opposite directions:
\[
x_i^{(\alpha)} = x_i - \alpha g,
\qquad
t_c^{(\alpha)} = t_c + \alpha g .
\]
Using an inner-product score, the corrected score at setting \(k=\alpha\) is
\[
\begin{aligned}
s_{ic}^{(\alpha)}
&=
\langle x_i-\alpha g, t_c+\alpha g\rangle \\
&=
\langle x_i,t_c\rangle
+
\alpha\langle x_i,g\rangle
-
\alpha\langle g,t_c\rangle
-
\alpha^2\|g\|^2 .
\end{aligned}
\]
For a fixed image \(i\), only the term \(-\alpha\langle g,t_c\rangle\) depends on the class \(c\).
Thus, before normalization, Linear correction induces a class-wise score bias.
For positive correction strength, we define this bias as
\begin{equation}
b_c = -\langle g,t_c\rangle .
\label{eq:gap_bias}
\end{equation}
Although our experiments use L2-normalized embeddings and cosine similarity after correction, the unnormalized term \(b_c\) provides a simple analytic proxy for the class-wise score bias induced by the gap direction.
Under this proxy, classes with larger \(b_c\) are expected to receive larger correction-induced score gains, which can unevenly change class rankings and favor certain class prototypes.
This provides a possible mechanism by which corrected predictions concentrate on a small subset of classes, forming prediction hubs.

We test this mechanism under Linear over-correction at \(\alpha=0.5\).
For each class \(c\), we compute three class-level quantities:
\(m_c^{(\alpha)}\), the total prediction count after correction;
\(\Delta m_c^{(\alpha)}\), its increase from the original model; and
\(e_c^{(\alpha)}\), the number of correct-to-wrong transitions redirected to class \(c\).
We then compute within-dataset Spearman correlations between \(b_c\) and these quantities to test whether the gap-induced bias predicts hub formation.

Table~\ref{tab:linear_bias_hub_corr} shows that the gap-induced bias is positively correlated with all three quantities across all datasets.
Classes favored by the correction bias tend to receive more predictions after correction, gain prediction count relative to the original model, and absorb more correct-to-wrong transitions.
These results indicate that prediction hubs under Linear over-correction are not arbitrary.
They are partly predictable from the class-wise score bias induced by the gap vector.

\section{Experiments: Prediction-Level Hubness and Accuracy Degradation}
\label{sec:experiments}

This section examines whether accuracy degradation under modality gap reduction is associated with prediction-level hubness.
We first describe the setup and metrics, then analyze the relationship between predicted-class concentration and accuracy, examine transition-level evidence, and finally test the proposed bias mechanism interventionally.
As in Section~\ref{sec:mechanism}, these analyses carry different evidential weight: the cross-method results (Sections~\ref{subsec:prediction_concentration_accuracy}--\ref{subsec:transition_evidence}) are \textbf{correlational}---for training-based profiles, prediction concentration is a diagnostic, not an established mechanism---whereas the intervention in Section~\ref{subsec:hub_suppression} is, like Section~\ref{subsec:linear_bias_hubs}, \textbf{mechanistic but Linear-specific}.

\subsection{Experimental Setup and Metrics}
\label{subsec:experimental_setup}

\begin{table*}[t]
\centering
\small
\setlength{\tabcolsep}{4pt}
\resizebox{\textwidth}{!}{
\begin{tabular}{lcccccccccccc}
\toprule
Method
& Cal101 & C10 & C100 & DTD & Euro & Aircraft & Flw102 & Food & IN & Pets
& Avg. & Exp. \\
\midrule
Linear
& -0.819 & -0.998 & -0.997 & -0.996 & -0.725 & -0.766 & -0.934 & -0.995 & -1.000 & -0.993
& -0.922 & 10/10 \\
CLIPRefine
& 0.799 & -0.923 & -0.913 & -0.268 & -0.630 & -0.431 & -0.122 & -0.900 & -0.711 & -0.750
& -0.485 & 9/10 \\
\bottomrule
\end{tabular}
}
\caption{
Within-dataset Spearman correlations between accuracy and Predicted-Class Gini for Linear correction and CLIPRefine.
``Exp.'' indicates the number of datasets with the expected negative correlation direction.
Section~\ref{subsec:reporting} explains the positive Caltech101 correlation under CLIPRefine.
}
\label{tab:within_dataset_gini_corr}
\end{table*}

\paragraph{Datasets and Correction Profiles}
We evaluate zero-shot image classification on 10 datasets:
Caltech101, CIFAR-10, CIFAR-100, DTD, EuroSAT, FGVC-Aircraft, Flowers102, Food-101, ImageNet-1K, and Oxford-IIIT Pet.~%
\footnote{\url{https://docs.pytorch.org/vision/main/datasets.html}},%
\footnote{\url{https://huggingface.co/datasets/ILSVRC/imagenet-1k}}
For each image, the predicted class is the class whose text prototype has the highest similarity with the image embedding.

We compare three correction profiles:
Linear correction~\cite{modality-gap}, CLIPRefine~\cite{CLIPRefine}, and AlignCLIP~\cite{AlignCLIP}.
For Linear correction and CLIPRefine, we use CLIP ViT-B/32~\cite{CLIP} as the base model.
For AlignCLIP, we use the public AlignCLIP checkpoint with its original ViT-B/16 backbone.
~\footnote{\url{https://huggingface.co/sarahESL/AlignCLIP}}
Linear correction provides a controlled post-hoc intervention, while CLIPRefine and AlignCLIP represent additional-training-based and pre-training-based correction profiles, respectively.

AlignCLIP is supplementary throughout and not part of the controlled comparison: its only public checkpoint uses a different backbone (ViT-B/16 versus ViT-B/32) with far fewer comparison points, so we report its results only as supporting evidence (Appendix~\ref{app:alignclip_results}).

For Linear correction, we sweep the correction strength \(\alpha\).
For learning-based corrections, we evaluate available checkpoints.
Unless otherwise stated, zero-shot scores are computed using cosine similarity between normalized image embeddings and text prototypes.
~\footnote{Further implementation details, including prompt templates, correction strengths, checkpoints, and backbone differences, are provided in Appendix~\ref{app:additional_experimental_details}.}

\paragraph{Metrics}
We report classification accuracy and prediction concentration.

For each evaluated correction step \(k\), let \(\hat{y}_i^{(k)}\) be the predicted label of sample \(i\).
We define the predicted-class count of class \(c\) as
\[
m_c^{(k)}
=
\sum_{i=1}^{N}
\mathbf{1}
[
\hat{y}_i^{(k)} = c
].
\]
This quantity counts how many samples are predicted as class \(c\).
Classes with large \(m_c^{(k)}\) are treated as prediction hubs.

To quantify prediction concentration, we apply the standard Gini coefficient~\cite{ceriani2012origins} to the predicted-class count vector
\(m^{(k)}=(m_1^{(k)},\ldots,m_C^{(k)})\), which we call Predicted-Class Gini.
A larger Predicted-Class Gini indicates that predictions are more concentrated on a smaller subset of classes.~\footnote{As a complementary concentration measure, we also report normalized prediction entropy in Appendix~\ref{app:hubness_details}.}
Let \(m_{(1)}^{(k)} \leq \cdots \leq m_{(C)}^{(k)}\) denote the sorted counts.
We define
\[
G^{(k)}
=
\frac{2\sum_{r=1}^{C} r\,m_{(r)}^{(k)}}
{C\sum_{r=1}^{C}m_{(r)}^{(k)}}
-
\frac{C+1}{C}.
\]

\paragraph{Transition Analysis}
As an additional diagnostic analysis, we analyze the destination distribution of prediction changes.
We decompose prediction changes into wrong-to-correct and correct-to-wrong transitions.
For correct-to-wrong transitions, we count how many previously correct samples are redirected to each destination class.
This analysis provides transition-level evidence for whether accuracy degradation is accompanied by many samples being redirected toward a small set of erroneous destination classes.~\footnote{The full transition-level destination curves are provided in Appendix~\ref{app:transition_full}.}



\begin{figure*}[t]
    \centering
    \includegraphics[keepaspectratio, scale=0.4]{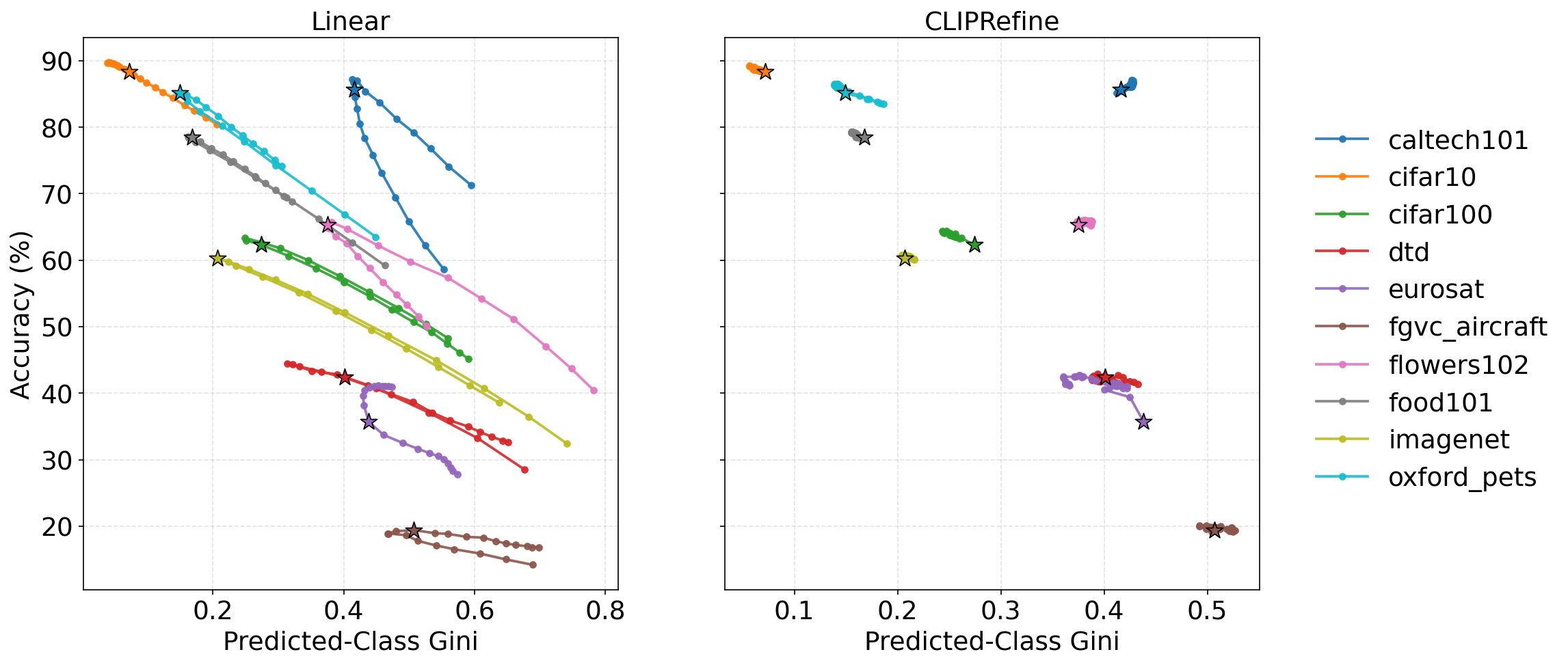}
    \caption{
    Predicted-Class Gini vs. accuracy for Linear correction and CLIPRefine.
    Stars indicate the base model.
    For both correction profiles, higher Predicted-Class Gini tends to correspond to lower accuracy, indicating that accuracy degradation is associated with increased prediction concentration.
    Pearson correlations are \(r=-0.778\) for Linear correction (\(n=210\)) and \(r=-0.710\) for CLIPRefine (\(n=310\)).
    }
    \label{fig:acc_gini}
\end{figure*}

\subsection{Prediction Concentration is Associated with Accuracy Degradation}
\label{subsec:prediction_concentration_accuracy}

We examine whether accuracy degradation under modality gap correction is associated with prediction-level hubness.
If prediction-level hubness reflects a failure mode of gap correction, then accuracy should decrease when the predicted-class distribution becomes more concentrated.
We focus on Linear correction and CLIPRefine in the main analysis, since they provide multiple correction settings under the same CLIP ViT-B/32 backbone.\footnote{AlignCLIP results are reported in Appendix~\ref{app:alignclip_results} as supplementary evidence because AlignCLIP uses a different backbone and has fewer comparison points in our setting.}

Figure~\ref{fig:acc_gini} shows the relationship between Predicted-Class Gini and accuracy for Linear correction and CLIPRefine.
For Linear correction, Predicted-Class Gini is negatively correlated with accuracy.
Moderate correction often reduces prediction concentration and improves accuracy, whereas stronger correction increases concentration and reduces accuracy.
Thus, the over-correction regime is not only a regime with smaller average modality gap, but also one in which predictions collapse toward fewer classes.

CLIPRefine shows a similar but weaker trend.
Although the range of Predicted-Class Gini is smaller than in Linear correction, higher Gini still tends to correspond to lower accuracy.
These results support the hypothesis that accuracy changes under modality gap correction are closely related to predicted-class concentration.\footnote{Appendix~\ref{app:hubness_details} reports normalized prediction entropy as a complementary metric and shows the corresponding trend.}

We further compute within-dataset Spearman correlations between accuracy and Predicted-Class Gini.
This analysis checks whether the observed relationship is driven only by cross-dataset differences.
For each dataset and method, we compute correlations across correction strengths or checkpoints, and then average the correlations across datasets.

Table~\ref{tab:within_dataset_gini_corr} shows that the expected relationship generally holds within individual datasets.
For Linear correction, all 10 datasets show the expected negative correlation between Predicted-Class Gini and accuracy.
CLIPRefine also shows the same direction in most datasets, although the relationship is weaker.
These results indicate that the concentration--accuracy relationship is not merely a pooled cross-dataset artifact.

Predicted-Class Gini is also not a redundant encoding of the correction strength itself: at every fixed non-zero \(\alpha\), the cross-dataset Spearman correlation between \(\Delta\mathrm{Acc}\) and \(\Delta G\) (changes from \(\alpha=0\)) is negative (mean \(-0.86\) for gap-reducing \(\alpha\); with \(n=10\) per point, 10 of 20 reach \(p<0.05\), and the claim rests on this directional consistency).
At the same strength, the datasets that concentrate more lose more accuracy---a vulnerability that \(\alpha\) alone does not express, as \(b_c\) predicts.

While these results establish a strong association between prediction concentration and accuracy degradation, they do not show how prediction changes are distributed across destination classes.
We therefore next examine whether harmful prediction changes concentrate on a small subset of destination classes.

\begin{table}[t]
\centering
\small
\setlength{\tabcolsep}{4pt}
\begin{tabular}{llrrr}
\toprule
Setting
& Transition
& Count
& Top-1
& Top-5 \\
\midrule
Linear-best
& C$\rightarrow$W
& 1,556
& 0.116
& 0.391 \\
Linear-best
& W$\rightarrow$C
& 3,487
& 0.187
& 0.396 \\
Linear-worst
& C$\rightarrow$W
& 33,804
& 0.294
& 0.672 \\
Linear-worst
& W$\rightarrow$C
& 5,238
& 0.272
& 0.585 \\
\midrule
CLIPRefine-best
& C$\rightarrow$W
& 3,547
& 0.156
& 0.460 \\
CLIPRefine-best
& W$\rightarrow$C
& 6,430
& 0.201
& 0.475 \\
CLIPRefine-worst
& C$\rightarrow$W
& 1,630
& 0.145
& 0.388 \\
CLIPRefine-worst
& W$\rightarrow$C
& 1,294
& 0.121
& 0.360 \\
\bottomrule
\end{tabular}
\caption{
Destination concentration of changed predictions.
C$\rightarrow$W denotes correct-to-wrong transitions, and W$\rightarrow$C denotes wrong-to-correct transitions.
Count is the total number of transitions over 10 datasets.
Top-1 and Top-5 report the mean fraction of transitions absorbed by the most frequent and top-five destination classes, respectively.
}
\label{tab:transition_destination_concentration}
\end{table}

\subsection{Transition-Level Evidence for Prediction Hubs}
\label{subsec:transition_evidence}

We next examine prediction changes directly through a transition-level analysis.
This analysis asks whether changed predictions concentrate on a small number of destination classes, complementing prior work on embedding-space and retrieval-space hubness~\citep{hubs-in-space,zero-shot-hub,balance-act,NeighborRetr,deguchi-etal-2026-one}.
We decompose changed predictions into wrong-to-correct and correct-to-wrong transitions and measure how concentrated their destination classes are.
If prediction hubs mediate accuracy degradation, correct-to-wrong transitions should become frequent and concentrate on a small subset of destination classes.

Table~\ref{tab:transition_destination_concentration} summarizes this destination concentration.
Here, ``best'' and ``worst'' denote the highest- and lowest-accuracy settings within each correction profile.
For Linear correction, the best setting has relatively few correct-to-wrong transitions, while the worst setting shows a large increase in harmful transitions.
In the Linear-worst setting, correct-to-wrong transitions rise to 33,804, and the top five destination classes absorb 67.2\% of these transitions on average.
This indicates that many previously correct samples are redirected into a small set of erroneous destination classes.

CLIPRefine shows a more balanced pattern.
Its best setting has more wrong-to-correct than correct-to-wrong transitions, and its worst setting has much lower destination concentration than Linear-worst.
For example, the top-five fraction for correct-to-wrong transitions is 0.388 in CLIPRefine-worst, compared with 0.672 in Linear-worst.
This is consistent with CLIPRefine's narrower range of prediction concentration and more stable accuracy profile.~\footnote{The full transition-level destination curves are provided in Appendix~\ref{app:transition_full}.}
These results support the view that prediction hubs can mediate accuracy degradation, but leave their source unresolved.
We next test whether the gap-induced class-wise bias \(b_c\), identified in Section~\ref{sec:mechanism}, accounts for these hubs through a score-space intervention.

\subsection{Intervening on Gap-Induced Bias}
\label{subsec:hub_suppression}

To assess the role of the gap-induced bias in hub formation, we introduce a score-space intervention.
In Section~\ref{sec:mechanism}, we showed that Linear correction induces the class-wise score bias \(b_c\), which favors classes with larger bias values.

We implement this intervention by modifying the corrected score as\footnote{
In implementation, this score-space intervention is applied on the CLIP logit scale: both \(s^{(\alpha)}_{i,c}\) and the subtracted bias term \(\alpha b_c\) are multiplied by the same factor 100.
This common positive scaling does not affect the argmax predictions.
}
\begin{equation}
s'_{i,c}
=
s^{(\alpha)}_{i,c}
-
\lambda \alpha b_c .
\label{eq:bias_subtraction}
\end{equation}
Here, \(s^{(\alpha)}_{i,c}\) is the score after Linear correction with strength \(\alpha\), and \(\lambda\) controls the intervention strength.
Because the first-order class-wise bias induced by Linear correction is \(\alpha b_c\), setting \(\lambda=1\) approximately cancels this bias.
Predictions after the intervention are obtained by
\[
\hat{y}'_i = \arg\max_c s'_{i,c}.
\]

We apply this intervention to the over-correction setting \(\alpha=0.5\), where prediction concentration is strongest.
For each dataset, we compare the pre-intervention point \((G^{(\alpha)}, \mathrm{Acc}^{(\alpha)})\) with the post-intervention point \((G', \mathrm{Acc}')\), where \(G\) denotes Predicted-Class Gini.
If the gap-induced bias contributes to hub formation, the intervention should move points toward lower prediction concentration and higher accuracy.

\begin{figure}[t]
    \centering
    \includegraphics[keepaspectratio, scale=0.35]{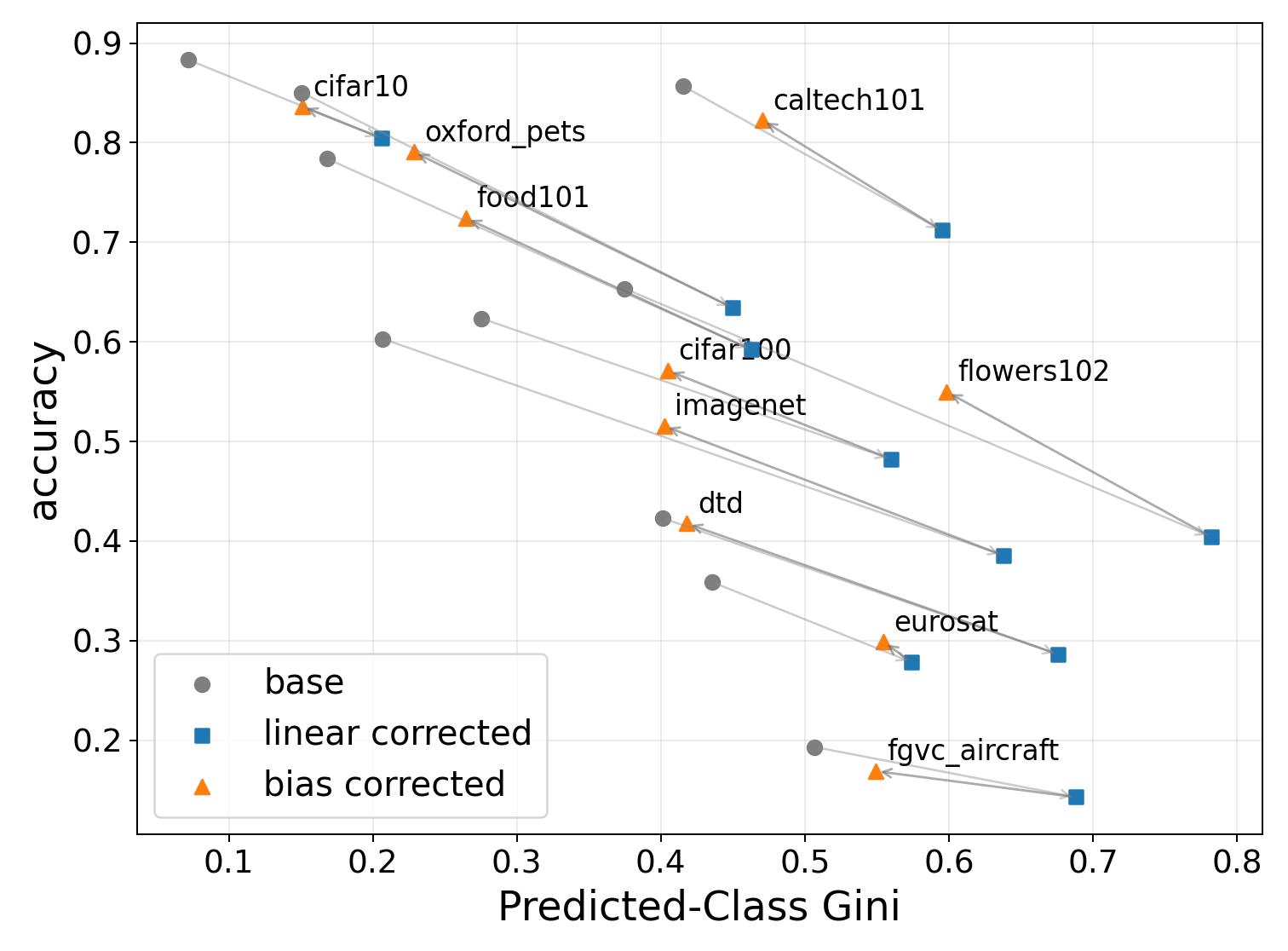}
    \caption{
    Effect of gap-bias subtraction after Linear over-correction.
    Gray circles denote the base model, blue squares denote Linear correction at \(\alpha=0.5\), and orange triangles denote the scores after subtracting the gap-induced class-wise bias with \(\lambda=1\).
    The markers illustrate the sequence from the base model to Linear over-correction and then to bias subtraction, which reduces Predicted-Class Gini and recovers the accuracy lost under over-correction.
    }
    \label{fig:gap_bias_suppression}
\end{figure}

Figure~\ref{fig:gap_bias_suppression} shows that subtracting the gap-induced class-wise bias consistently reduces Predicted-Class Gini across datasets.
In most cases, this reduction in prediction concentration partially recovers the accuracy lost under Linear over-correction, moving performance closer to the base model.
This result provides interventional evidence for the proposed mechanism: the class-wise bias derived from the modality gap is not only correlated with prediction hubs, but contributes to their formation.
At the same time, the intervention does not claim to be a practical correction method; rather, it serves as a diagnostic test of the bias-driven hubness mechanism.%
\footnote{See Appendix~\ref{app:alpha_sweep} for additional results sweeping both the correction strength $\alpha$ and the bias-subtraction strength $\lambda$.}

\subsection{Interpreting and Mitigating Prediction Concentration}
\label{subsec:reporting}
\label{subsec:mitigation}

Predicted-Class Gini should be read relative to the label distribution of the evaluation set: on an imbalanced set even a perfect classifier produces an imbalanced prediction distribution, so concentration mixes a benign component (fitting the label prior) with the failure mode studied here (correction-induced excess concentration).
The single anomaly in Table~\ref{tab:within_dataset_gini_corr} is the benign case: Caltech101 (label Gini 0.402, by far our most imbalanced) gains accuracy under CLIPRefine (85.70\% to 87.09\%) while Predicted-Class Gini stays near the label Gini (0.416 to 0.427), and dividing each predicted-class count by its ground-truth frequency reverses the correlation from \(+0.799\) to \(-0.979\) while leaving balanced datasets essentially unchanged (Appendix~\ref{app:generic_config}).
Linear over-correction is the excess case: Gini reaches 0.595, far above the label Gini, with the correlation staying negative (\(-0.819\)).
We therefore recommend reporting the average modality gap, accuracy, Predicted-Class Gini, and normalized prediction entropy relative to the uncorrected base model, together with the label Gini; we propose no universal absolute-Gini threshold, since its range depends on the label distribution and class count.

Two mitigation options follow, and neither is a solution.
First, modality-wise centering---the embedding-space counterpart of the score-space subtraction in Section~\ref{subsec:hub_suppression}---erases the \(\alpha\) dependence almost entirely (Appendix~\ref{app:centering}); this further confirms the \(b_c\) mechanism but is diagnostic rather than practical, as it removes the very quantity under study and shifts base accuracy in a dataset-dependent way.
Second, hubness-aware scoring: CSLS~\cite{CSLS} roughly halves the over-correction penalty but leaves the concentration--accuracy relationship intact on all 10 datasets (Appendix~\ref{app:csls_details}), so prediction-level hubness is not an artifact of raw cosine similarity.
Making gap reduction itself hubness-aware is left to future work.

\section{Discussion}
\label{sec:discussion}

\paragraph{What Makes Gap Reduction Useful?}

Our results suggest that modality gap reduction is useful only when it preserves the decision structure needed for downstream prediction.
Linear correction reduces average image--text displacement, but can simultaneously distort downstream decision geometry through uneven class-wise score shifts.
When these shifts favor a small subset of class prototypes, predictions become concentrated into prediction hubs and accuracy deteriorates.
In contrast, CLIPRefine reduces the gap without inducing this concentration, suggesting that a learned correction can preserve the decision structure that Linear correction distorts.
The explicit bias decomposition in Section \ref{sec:mechanism} is specific to Linear correction, which we use as an analytically tractable example of how modality-gap reduction can distort downstream prediction structure.

\paragraph{A Candidate Geometric Correlate}

What geometric property separates Linear correction, which concentrates predictions as it reduces the gap, from CLIPRefine, which does not?
Not prototype arrangement: both profiles preserve the prototype--prototype angular structure almost exactly, even under over-correction.
A better candidate is joint image--text uniformity~\cite{contrastive-learning}: under Linear correction it improves only at small correction strengths and then deteriorates just past the accuracy peak, tracking accuracy and Gini on all 10 datasets, whereas CLIPRefine shrinks the gap while \emph{improving} uniformity, with Gini essentially unchanged (Appendices~\ref{app:uniformity_details} and~\ref{app:uniformity_results}).
We present this as a candidate correlate, not a causal explanation (the correlation reverses on EuroSAT under the generic template; Appendix~\ref{app:generic_config}); whether uniformity can be constrained during correction or post-training is future work.

\paragraph{Is the Average Modality Gap the Right Quantity?}

Following prior work on modality gap estimation~\cite{modality-gap}, we measure the modality gap as the squared Euclidean distance between mean image and text embeddings.
However, CLIP zero-shot classification is typically based on cosine similarity between normalized embeddings.
Thus, the metric used to quantify the gap is not identical to the scoring rule used for prediction.
This mismatch suggests that future work should consider gap measures that are more directly tied to the geometry of downstream decision rules.

\paragraph{Rethinking the Goal of Modality Gap Reduction}

Our results raise a broader question: if reducing the modality gap does not necessarily improve performance, what should gap reduction aim to achieve?
One answer is that gap reduction should not be treated as an end in itself.
Rather, it should aim to improve cross-modal alignment without distorting the output distribution.
This suggests that modality gap correction should be evaluated not only by alignment metrics, but also by output-space diagnostics such as predicted-class concentration.

\section{Conclusion}
\label{sec:conclusion}

We studied why reducing the modality gap does not necessarily improve downstream zero-shot image classification.
We showed that gap correction can unevenly change class-wise decision margins and induce \textit{prediction-level hubness}, where predictions concentrate on a small subset of class prototypes.
Across correction profiles and datasets, accuracy degradation is associated with increased prediction-level hubness, and a score-space intervention on the gap-induced bias reduces this concentration under Linear over-correction.
These findings suggest that modality gap correction should not be evaluated solely by average image--text alignment, but also by how it affects downstream decision structure and output distributions.

\section*{Limitations}
\label{sec:limitations}

Our claims are scoped to zero-shot image classification, where predictions over a fixed set of class-text prototypes make predicted-class concentration directly observable and make accuracy a function of exactly the quantity our analysis decomposes; we do not claim that they carry over to cross-modal retrieval.
The bias algebra itself would transfer to retrieval with a fixed gallery (each gallery item acquires the same query-independent bias \(b_c=-\langle g, t_c\rangle\), and the derivation of Equation~\ref{eq:gap_bias} is unchanged), but retrieval returns ranked lists rather than a single label, so both the concentration metric and the accuracy analysis would have to be reformulated---plausibly in terms of top-\(k\) occupancy across queries---which we have not done.
Because the failure mode we identify is specific to a fixed prototype set competing for an argmax, gap reduction may well be genuinely beneficial for retrieval; our results are not evidence against it, and the same caution applies to captioning and visual question answering, where the output space is structured differently again.

Our notion of hubness is prediction-level rather than embedding-level.
A prediction hub in this work is a class label that receives a disproportionate number of final predictions after gap correction.
This differs from conventional nearest-neighbor hubness in embedding space~\citep{hubs-in-space}, and we use the prediction-level sense consistently throughout.
Our results should therefore be understood as an analysis of how gap correction affects classifier outputs, rather than as a complete account of the underlying representation geometry.

Our evidence is not uniformly mechanistic, as set out in Section~\ref{sec:mechanism}: for Linear correction we have a closed-form class-wise bias, a prediction of which classes become hubs, and an intervention that cancels the bias and reverses the effect, whereas for CLIPRefine---the only training-based profile with enough comparison points to analyze---prediction concentration is diagnostic rather than causal.
CLIPRefine~\cite{CLIPRefine} and AlignCLIP~\cite{AlignCLIP} change the encoders themselves and may reflect effects beyond average gap reduction, so our mechanistic account covers Linear correction specifically, not learning-based correction in general; the joint uniformity result in Section~\ref{sec:discussion} sits at the same correlational level.

AlignCLIP is evaluated with fewer comparison points and on a different backbone, so we treat it as supplementary evidence throughout rather than as part of the controlled comparison.
Our controlled results use a single backbone, ViT-B/32; Appendix~\ref{app:model_scale} finds the same phenomenon for the Linear-correction sweep on ViT-B/16 and ViT-L/14, but the training-based profiles are only available to us at one scale.
Future work should test hubness-aware correction objectives that reduce prediction concentration while preserving the benefits of modality alignment, and extend the analysis to output spaces beyond a fixed label set.


\section*{Acknowledgements}
This work was supported by JST PRESTO, Japan, Grant Number JPMJPR2366.

The authors used ChatGPT to support research discussion, idea organization, and language editing.
In particular, generative assistance was used to discuss and clarify author-developed claims and experimental designs, improve the clarity of exposition, and draft or revise low-novelty text based on author-provided content.
All research ideas, experimental designs, analyses, interpretations, citations, and final text were reviewed, verified, and approved by the authors, who take full responsibility for the submitted content.


\bibliography{custom}

\appendix

\section{Additional Experimental Details}
\label{app:additional_experimental_details}

This appendix provides implementation details shared across the experiments in the main text.
These details apply to the gap--accuracy comparison in Figure~\ref{fig:rawdual} as well as to the later prediction-level hubness analyses.
We describe the evaluation datasets, the prompt templates used to construct class-text prototypes, the implementation of Linear correction~\cite{modality-gap}, the computation of the modality gap, the backbone and checkpoint settings for the learning-based correction profiles~\cite{CLIPRefine,AlignCLIP}, the model-loading, preprocessing, and evaluation settings, and the artifact licenses and terms of use.

\subsection{Datasets}
\label{app:datasets}

We evaluate zero-shot image classification on ten standard image-classification datasets: Caltech101, CIFAR-10, CIFAR-100, DTD, EuroSAT, FGVC-Aircraft, Flowers102, Food-101, ImageNet-1K, and Oxford-IIIT Pet.
For datasets with standard evaluation splits, we use the test or validation split commonly used for evaluation. For datasets without such a split in the dataset interface used in our experiments, such as Caltech101 and EuroSAT, we evaluate on the full dataset.
No new images or labels are collected in this work.
The datasets are used only for aggregate zero-shot evaluation and for computing derived prediction-level statistics such as predicted-class count, Predicted-Class Gini, normalized prediction entropy, and transition counts.

\begin{table}[h]
\centering
\small
\begin{tabular}{lrr}
\toprule
\textbf{Dataset} & \textbf{\#Classes} & \textbf{Evaluation set} \\
\midrule
Caltech101 & 101 & Full dataset \\
CIFAR-10 & 10 & Test split \\
CIFAR-100 & 100 & Test split \\
DTD & 47 & Test split \\
EuroSAT & 10 & Full dataset \\
FGVC-Aircraft & 100 & Test split \\
Flowers102 & 102 & Test split \\
Food-101 & 101 & Test split \\
ImageNet-1K & 1000 & Validation split \\
Oxford-IIIT Pet & 37 & Test split \\
\bottomrule
\end{tabular}
\caption{Datasets used for zero-shot image-classification evaluation. For datasets without an official evaluation split in the dataset interface used in our experiments, we evaluate on the full dataset. We do not collect or redistribute any original images or labels.}
\label{tab:app_datasets}
\end{table}

\subsection{Prompt Templates}
\label{app:prompt_templates}

For zero-shot classification with CLIP~\cite{CLIP}, we construct one text prototype for each class using a dataset-specific prompt template.
Table~\ref{tab:app_prompt_templates} lists the templates, which are used in all experiments reported in this paper unless stated otherwise.
The five datasets whose entry in Table~\ref{tab:app_prompt_templates} is the generic \texttt{a photo of a \{class name\}.} are of course identical under either choice; for the other five, Appendix~\ref{app:generic_config} reports the full set of results obtained with the generic template applied uniformly, as a robustness check.
All conclusions in this paper hold under both configurations.

\begin{table*}[t]
\centering
\small
\begin{tabular}{ll}
\toprule
Dataset & Prompt template \\
\midrule
CIFAR-10, CIFAR-100, Caltech101, ImageNet-1K, Oxford-IIIT Pet
& \texttt{a photo of a \{class name\}.} \\
DTD
& \texttt{a photo of a \{class name\} texture.} \\
EuroSAT
& \texttt{a satellite photo of a \{class name\}.} \\
FGVC-Aircraft
& \texttt{a photo of a \{class name\} aircraft.} \\
Flowers102
& \texttt{a photo of a \{class name\} flower.} \\
Food-101
& \texttt{a photo of \{class name\}.} \\
\bottomrule
\end{tabular}
\caption{
Prompt templates used for constructing class-text prototypes.
}
\label{tab:app_prompt_templates}
\end{table*}

\begin{table*}[t]
\centering
\small
\begin{tabularx}{\textwidth}{p{0.25\textwidth} p{0.35\textwidth} X}
\toprule
\textbf{Artifact} & \textbf{Access / Original Source} & \textbf{License / Terms} \\
\midrule
CLIP ViT-B/32
& OpenAI CLIP release
& MIT License \\

CLIPRefine
& Official public implementation/checkpoints of Yamaguchi et al. (2025)
& NTT Software License Agreement for Evaluation \\

AlignCLIP
& Public Hugging Face checkpoint by Eslami and de Melo (2025)
& CC-BY-NC-ND-4.0 according to the model card \\

Standard image-classification datasets
& Caltech101, CIFAR-10/100, DTD, EuroSAT, FGVC-Aircraft, Flowers102, Food-101, ImageNet-1K, and Oxford-IIIT Pet
& Respective licenses or terms of the original dataset releases \\
\bottomrule
\end{tabularx}
\caption{Artifacts used in this work and their licenses or terms of use. All artifacts are used only for research evaluation, and we do not redistribute the original datasets, images, software, or model checkpoints.}
\label{tab:artifact-licenses}
\end{table*}

\subsection{Linear Correction Implementation}
\label{app:linear_correction_details}

For Linear correction, we follow the embedding-shift formulation of Liang et al.~\cite{modality-gap}.
Given image embeddings \(x_i\) and text prototypes \(t_j\), we estimate the gap vector as
\[
g
=
\frac{1}{N}\sum_{i=1}^{N}x_i
-
\frac{1}{N}\sum_{j=1}^{N}t_j .
\]
For correction strength \(\alpha\), image and text embeddings are shifted in opposite directions:
\[
x_i^{*}=x_i-\alpha g,
\qquad
t_c^{*}=t_c+\alpha g.
\]
We then L2-normalize the shifted embeddings before scoring:
\[
\tilde{x}_i
=
\frac{x_i^{*}}{\|x_i^{*}\|_2},
\qquad
\tilde{t}_c
=
\frac{t_c^{*}}{\|t_c^{*}\|_2}.
\]
The predicted class is obtained by
\[
\hat{y}_i
=
\arg\max_c
\tilde{x}_i^\top \tilde{t}_c .
\]
The original CLIP prediction corresponds to \(\alpha=0\).
In our experiments, we sweep \(\alpha\) from \(-0.5\) to \(0.5\) with step size \(0.05\).
Increasing \(\alpha\) typically reduces the average modality gap, whereas decreasing \(\alpha\) expands it.
This setting allows us to continuously vary the degree of gap correction while keeping the pre-trained CLIP encoders fixed.

\subsection{Modality Gap Computation}
\label{app:modality_gap_details}

Following the gap-vector estimate used in Linear correction~\cite{modality-gap}, we compute the modality gap as the squared Euclidean distance between the mean image embedding and the mean text embedding:
\[
\mathrm{MG}
=
\left\|
\frac{1}{N}\sum_{i=1}^{N}x_i
-
\frac{1}{N}\sum_{j=1}^{N}t_j
\right\|_2^2 .
\]
For Linear correction, this quantity is computed after applying each correction strength \(\alpha\).
For learning-based correction profiles, it is computed at each evaluated checkpoint using the image and text representations produced by that checkpoint.
We do not apply an additional Linear correction to CLIPRefine or AlignCLIP when computing their checkpoint profiles.
The same definition is used for the modality-gap curves in Figure~\ref{fig:rawdual} and for the subsequent analyses.

\subsection{Joint Uniformity Computation}
\label{app:uniformity_details}

For the analysis in Section~\ref{sec:discussion}, we compute uniformity following \citet{contrastive-learning}.
Let \(Z\) be the joint set consisting of the corrected, L2-normalized image embeddings together with each sample's corrected ground-truth text prototype (one prototype instance per sample).
Uniformity is
\[
U
=
\frac{1}{|Z|^2}
\sum_{z_i, z_j \in Z}
\exp\!\left(-2\,\|z_i - z_j\|_2^2\right),
\]
where smaller values indicate a more uniformly spread joint distribution.
For computational efficiency, \(Z\) is subsampled to 2{,}000 points when larger; diagonal pairs are included.
For Linear correction, \(U\) is computed at each correction strength \(\alpha\); for CLIPRefine, at each checkpoint.
Because the joint set mixes both modalities, this quantity is sensitive to cross-modal collapse along the gap direction, unlike within-modality uniformity.

\subsection{Learning-Based Correction Profiles}
\label{app:learning_based_details}

We evaluate CLIPRefine~\cite{CLIPRefine} as an additional-training-based correction profile and AlignCLIP~\cite{AlignCLIP} as a pre-training-based correction profile.
For Linear correction~\cite{modality-gap} and CLIPRefine~\cite{CLIPRefine}, we use CLIP ViT-B/32~\cite{CLIP} as the base model.
For AlignCLIP~\cite{AlignCLIP}, we use the public AlignCLIP checkpoint with its original ViT-B/16 backbone.
For CLIPRefine, we evaluate the available public checkpoints in ascending checkpoint order, without applying an additional post-hoc Linear correction.
Because AlignCLIP uses a different backbone and provides fewer comparison points in our setting, we treat its results as supplementary evidence rather than a directly controlled comparison with Linear correction and CLIPRefine.
This distinction applies both to the gap--accuracy comparison in Figure~\ref{fig:rawdual} and to the prediction-concentration analyses in the main text.
\subsection{Implementation Details}
\label{app:software_implementation}

We used public model implementations and standard dataset interfaces
throughout the evaluation pipeline. For CLIP-based experiments, the
OpenAI CLIP model was loaded using \texttt{clip.load(model,
jit=False)}. For AlignCLIP experiments, we used the public AlignCLIP
checkpoint together with the associated model-construction function,
preprocessing transforms, and tokenizer. Unless otherwise specified,
image preprocessing followed the default preprocessing pipeline
provided by the corresponding model implementation.

Datasets were loaded with \texttt{torchvision.datasets} when
available. Evaluation was performed using a PyTorch
\texttt{DataLoader} with \texttt{batch\_size=256},
\texttt{num\_workers=4}, \texttt{shuffle=False}, and
\texttt{pin\_memory=True} by default. Zero-shot text prototypes were
constructed from the dataset-specific prompt templates listed in
Table~\ref{tab:app_prompt_templates}. Image and text embeddings were
L2-normalized before cosine-similarity scoring.

Predicted-Class Gini, normalized prediction entropy, transition
counts, Linear correction, CSLS scoring, and the bias-subtraction
intervention were implemented with custom PyTorch/NumPy code. For
CSLS, we used \(k=10\) in all experiments. Pearson and Spearman
correlations reported in the main analyses were computed using
\texttt{scipy.stats.pearsonr} and \texttt{scipy.stats.spearmanr},
respectively.

\begin{table*}[t]
\centering
\small
\setlength{\tabcolsep}{4pt}
\resizebox{\textwidth}{!}{
\begin{tabular}{lcccccccccccc}
\toprule
Method
& Cal101 & C10 & C100 & DTD & Euro & Aircraft & Flw102 & Food & IN & Pets
& Avg. & Exp. \\
\midrule
Linear
& 0.432 & 0.998 & 0.990 & 0.998 & 0.740 & 0.759 & 0.934 & 0.995 & 0.987 & 0.980
& 0.881 & 10/10 \\
CLIPRefine
& -0.972 & 0.951 & 0.904 & 0.181 & 0.602 & 0.398 & 0.227 & 0.909 & 0.654 & 0.853
& 0.471 & 9/10 \\
\bottomrule
\end{tabular}
}
\caption{
Within-dataset Spearman correlations between accuracy and normalized prediction entropy for Linear correction and CLIPRefine.
``Exp.'' indicates the number of datasets with the expected positive correlation direction.
}
\label{tab:within_dataset_entropy_corr}
\end{table*}

\begin{figure*}[t]
    \centering
    \includegraphics[width=\textwidth]{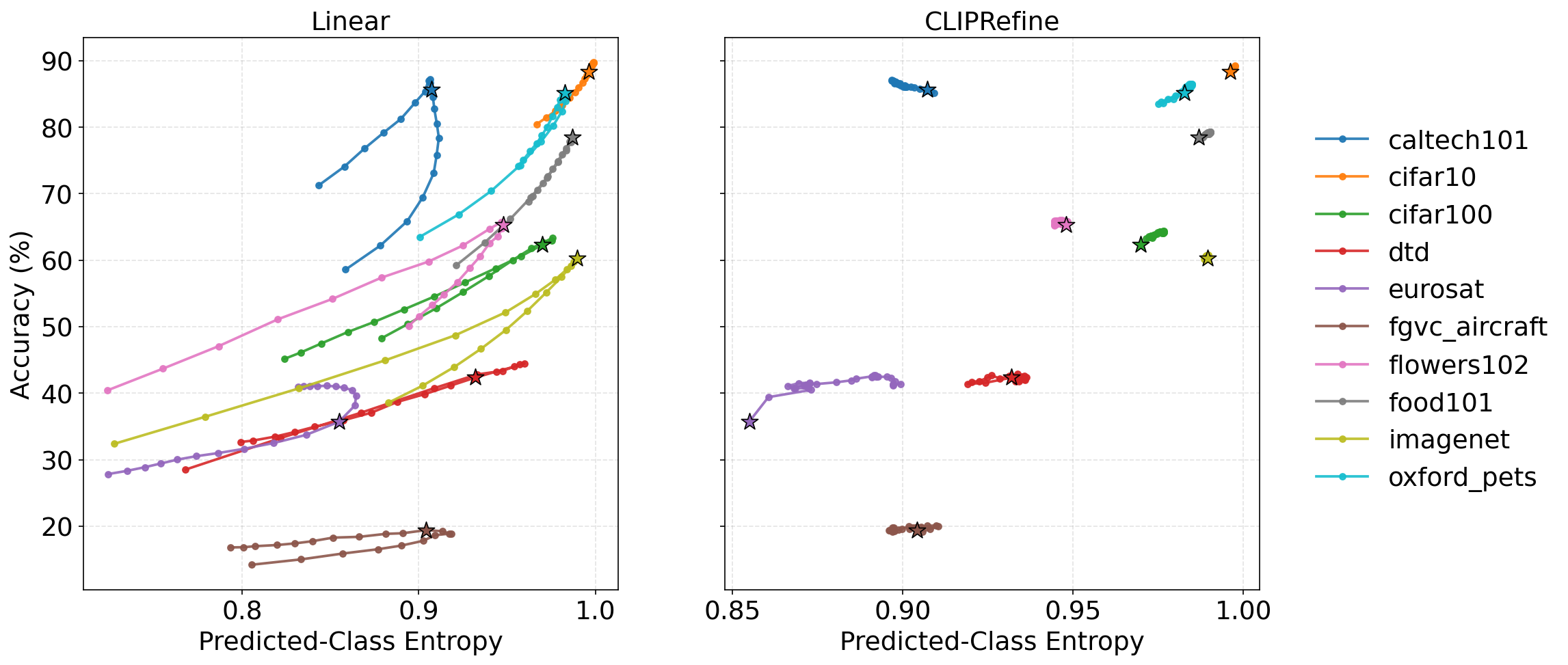}
    \caption{
    Normalized prediction entropy vs. accuracy for Linear correction and CLIPRefine.
    Stars indicate the base model.
    In both settings, higher normalized prediction entropy is associated with higher accuracy
    (Pearson's \(r=0.716\), \(n=210\) for Linear; \(r=0.585\), \(n=310\) for CLIPRefine).
    }
    \label{fig:acc_entropy}
\end{figure*}

\subsection{Artifact Licenses and Terms of Use}

Table~\ref{tab:artifact-licenses} summarizes the pretrained models,
checkpoints, software, and standard image-classification datasets used
in this work, together with their access sources and license or usage
terms. We use these publicly available artifacts for research
evaluation. Some datasets were accessed through standard dataset
interfaces such as \texttt{torchvision}, but their licenses and terms
are those of the original dataset releases rather than
\texttt{torchvision} itself. We follow the access conditions of the
original releases and do not redistribute the original datasets,
images, software, or pretrained checkpoints. Our use of these
artifacts is limited to research evaluation of zero-shot image
classification and modality-gap correction, which is consistent with
their research or evaluation-oriented access conditions.


\section{Prediction-Level Hubness Metrics and Complementary Results}
\label{app:hubness_details}

This appendix provides definitions and complementary analyses for the prediction-level hubness metrics used in the main text.
While the main analysis focuses on Predicted-Class Gini as the primary measure of prediction concentration, we also report normalized prediction entropy as a complementary measure of prediction diversity.
We further define the predicted-class count and transition-level destination concentration used to analyze how predictions concentrate across classes.
Because all of these metrics are computed from the argmax predictions, they are exactly invariant to any positive global rescaling of the logits: sweeping the inverse temperature from 10 to 500 at \(\alpha \in \{0, 0.25, 0.5\}\) leaves the predictions, and hence every metric below, unchanged on all 10 datasets.
Temperature acts uniformly across classes and cannot generate the class-dependent score shifts (such as \(b_c\)) required for hub formation, so the phenomena studied in this paper are not artifacts of the logit scale. 
All metrics are computed from final predicted labels rather than directly from embedding-space nearest-neighbor structure.

\subsection{Predicted-Class Count}
\label{app:predicted_class_mass}

For each correction point \(k\), let \(\hat{y}_i^{(k)}\) be the predicted label of sample \(i\).
The predicted-class count of class \(c\) is
\[
m_c^{(k)}
=
\sum_{i=1}^{N}
\mathbf{1}
[
\hat{y}_i^{(k)} = c
].
\]
This quantity counts how many samples are predicted as class \(c\).
Classes with large \(m_c^{(k)}\) are treated as prediction hubs.

\subsection{Predicted-Class Gini}
\label{app:hub_mass_gini}

We measure prediction concentration using the Gini coefficient over the predicted-class count vector
\[
m^{(k)}
=
(m_1^{(k)},\ldots,m_C^{(k)}).
\]
Let
\[
m_{(1)}^{(k)} \leq \cdots \leq m_{(C)}^{(k)}
\]
denote the sorted counts.
We define Predicted-Class Gini as
\[
G^{(k)}
=
\frac{2\sum_{r=1}^{C} r\,m_{(r)}^{(k)}}
{C\sum_{r=1}^{C}m_{(r)}^{(k)}}
-
\frac{C+1}{C}.
\]
A larger value indicates that predictions are more concentrated on a smaller subset of classes.

\begin{figure*}[t]
    \centering
    \includegraphics[keepaspectratio, width=\textwidth]{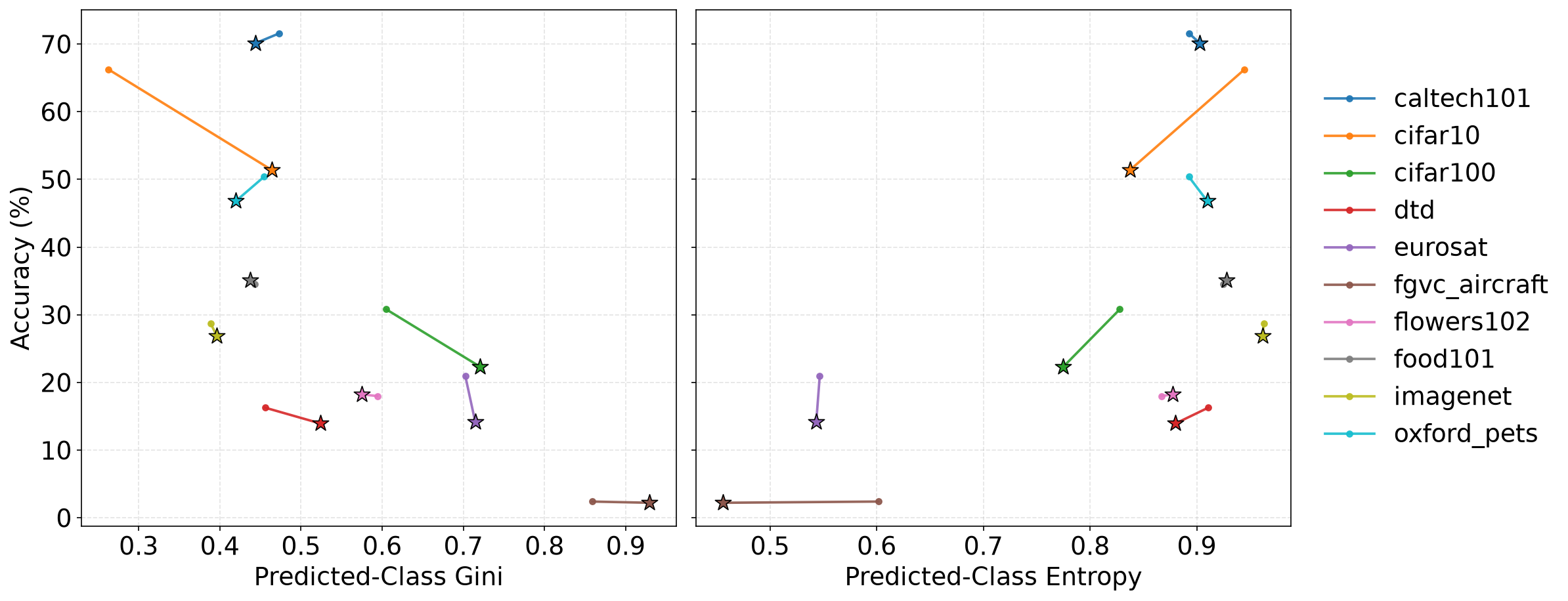}
    \caption{
    Predicted-Class Gini (left) and normalized prediction entropy (right) vs. accuracy for AlignCLIP.
    Stars indicate the base model.
    A negative correlation with Predicted-Class Gini (Pearson's \(r=-0.701\), \(n=20\)) and a positive correlation with entropy (\(r=0.556\), \(n=20\)) are observed, though the limited number of comparison points calls for cautious interpretation.
    }
    \label{fig:app_alignclip_corr}
\end{figure*}

\subsection{Normalized Prediction Entropy}
\label{app:normalized_prediction_entropy}

As a complementary concentration metric, we use normalized prediction entropy.
Let
\[
p_c^{(k)}
=
\frac{m_c^{(k)}}{\sum_{j=1}^{C}m_j^{(k)}} .
\]
We define
\[
H_{\mathrm{norm}}^{(k)}
=
-
\frac{
\sum_{c=1}^{C} p_c^{(k)}\log p_c^{(k)}
}
{\log C}.
\]
Terms with \(p_c^{(k)}=0\) are treated as zero.
A larger entropy value indicates that predictions are more evenly distributed across classes.
Thus, normalized entropy is expected to show the complementary trend to Predicted-Class Gini: if prediction concentration is associated with accuracy degradation, entropy should be positively correlated with accuracy.



Figure~\ref{fig:acc_entropy} shows that normalized prediction entropy is positively correlated with accuracy for both Linear correction and CLIPRefine.
This is the complementary pattern to the negative relationship between Predicted-Class Gini and accuracy reported in the main text.
The result indicates that accuracy degradation is associated with reduced diversity in the predicted-class distribution, not only with increased Gini concentration.


Table~\ref{tab:within_dataset_entropy_corr} further confirms the complementary entropy trend within individual datasets.
For Linear correction, all 10 datasets show the expected positive correlation between normalized entropy and accuracy.
For CLIPRefine, the expected direction holds in 9 out of 10 datasets, although the average correlation is weaker than for Linear correction.

\subsection{Transition-Level Destination Concentration}
\label{app:transition_destination_concentration}

To analyze where changed predictions go, we decompose prediction changes into wrong-to-correct and correct-to-wrong transitions.
For harmful correct-to-wrong transitions, the destination count of class \(c\) is
\[
e_c^{(k)}
=
\sum_i
\mathbf{1}
[
\hat{y}_i^{(0)} = y_i,\,
\hat{y}_i^{(k)} = c,\,
c \neq y_i
].
\]
If a small number of classes have large \(e_c^{(k)}\), then previously correct samples are being redirected toward a small set of erroneous destination classes.
This provides transition-level evidence for prediction-level hubness.

\section{Supplementary Results}
\label{app:supplementary_results}

This appendix collects supplementary results that extend the analyses in the main text: AlignCLIP results (Appendix~\ref{app:alignclip_results}), full transition-level destination curves (Appendix~\ref{app:transition_full}), the full sweep for the bias-subtraction intervention (Appendix~\ref{app:alpha_sweep}), and the prototype-structure and uniformity measurements behind the geometric-correlate discussion (Appendix~\ref{app:uniformity_results}).

\subsection{AlignCLIP Results}
\label{app:alignclip_results}

The main text focuses primarily on Linear correction~\cite{modality-gap} and CLIPRefine~\cite{CLIPRefine}, which provide more comparison points under comparable backbone settings.
Here, we report the corresponding prediction-concentration results for AlignCLIP~\cite{AlignCLIP}.

Figure~\ref{fig:app_alignclip_corr} shows that AlignCLIP follows the same qualitative trend as the main results:
higher predicted-class concentration corresponds to lower accuracy, while higher normalized entropy corresponds to higher accuracy.
Because the number of comparison points is limited, we use these results only as supporting evidence.


\subsection{Full Transition-Level Destination Curves}
\label{app:transition_full}

The main text summarizes the transition-level evidence for prediction-level hubness.
Here, we provide the full rank curves for all datasets.
This analysis complements prior work on embedding-space and retrieval-space hubness~\citep{hubs-in-space,zero-shot-hub,balance-act,NeighborRetr,deguchi-etal-2026-one} by examining whether final prediction changes concentrate on a small number of destination labels.

We decompose changed predictions into wrong-to-correct and correct-to-wrong transitions.
For each transition type, destination classes are sorted by transition count within each curve.
Thus, the curves compare concentration patterns rather than class identities.
A heavier head for correct-to-wrong transitions indicates that previously correct samples are redirected into a smaller number of erroneous destination classes.

\begin{figure*}[t]
    \centering
    \includegraphics[keepaspectratio, scale=0.49]{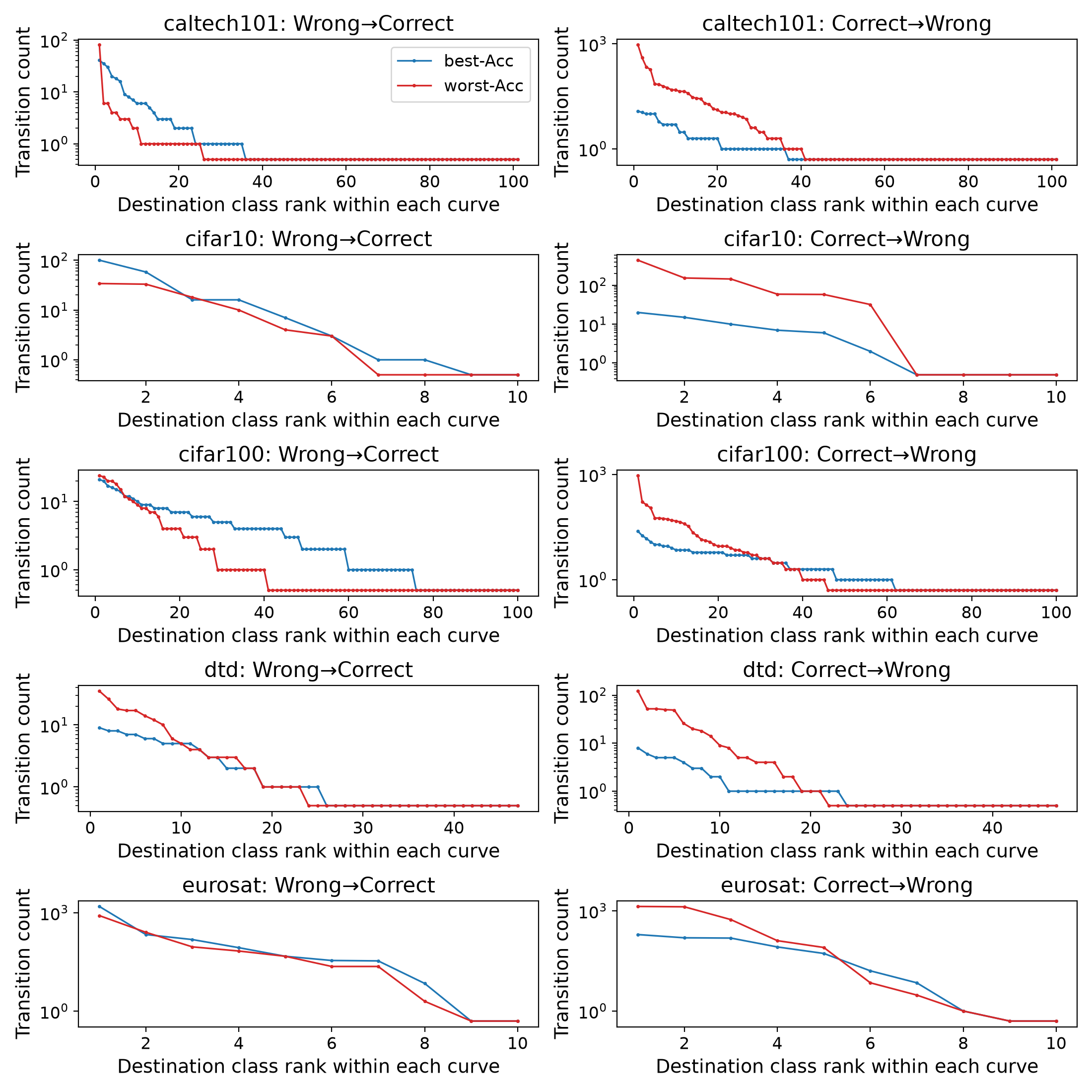}
    \caption{
    Rank curves of destination concentration for prediction transitions under Linear correction
    on the first five datasets: Caltech101, CIFAR-10, CIFAR-100, DTD, and EuroSAT.
    Destination classes are independently sorted by transition count within each curve.
    For visualization on the log-scale y-axis, destination classes with zero transitions are shown at 0.5.
    }
    \label{fig:app_prediction_destination_01_linear}
\end{figure*}

\begin{figure*}[t]
    \centering
    \includegraphics[keepaspectratio, scale=0.49]{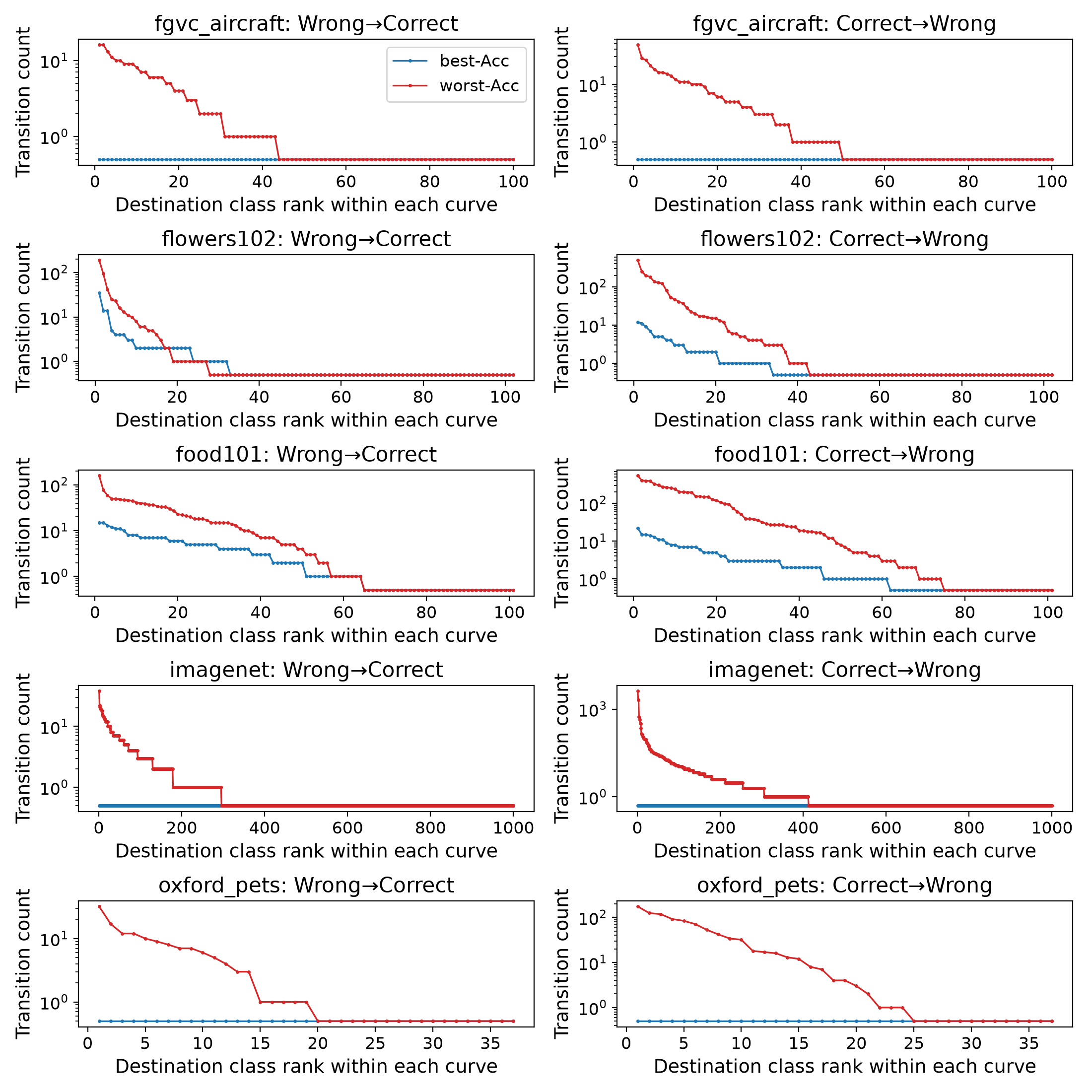}
    \caption{
    Rank curves of destination concentration for prediction transitions under Linear correction
    on the remaining five datasets: FGVC-Aircraft, Flowers102, Food-101, ImageNet-1K, and Oxford-IIIT Pet.
    Destination classes are independently sorted by transition count within each curve.
    For visualization on the log-scale y-axis, destination classes with zero transitions are shown at 0.5.
    }
    \label{fig:app_prediction_destination_02_linear}
\end{figure*}

Figures~\ref{fig:app_prediction_destination_01_linear} and~\ref{fig:app_prediction_destination_02_linear} show the full destination-rank curves for Linear correction~\cite{modality-gap}.
They provide the dataset-level transition patterns underlying the summary in Section~\ref{subsec:transition_evidence}.

\begin{figure*}[t]
    \centering
    \includegraphics[keepaspectratio, scale=0.49]{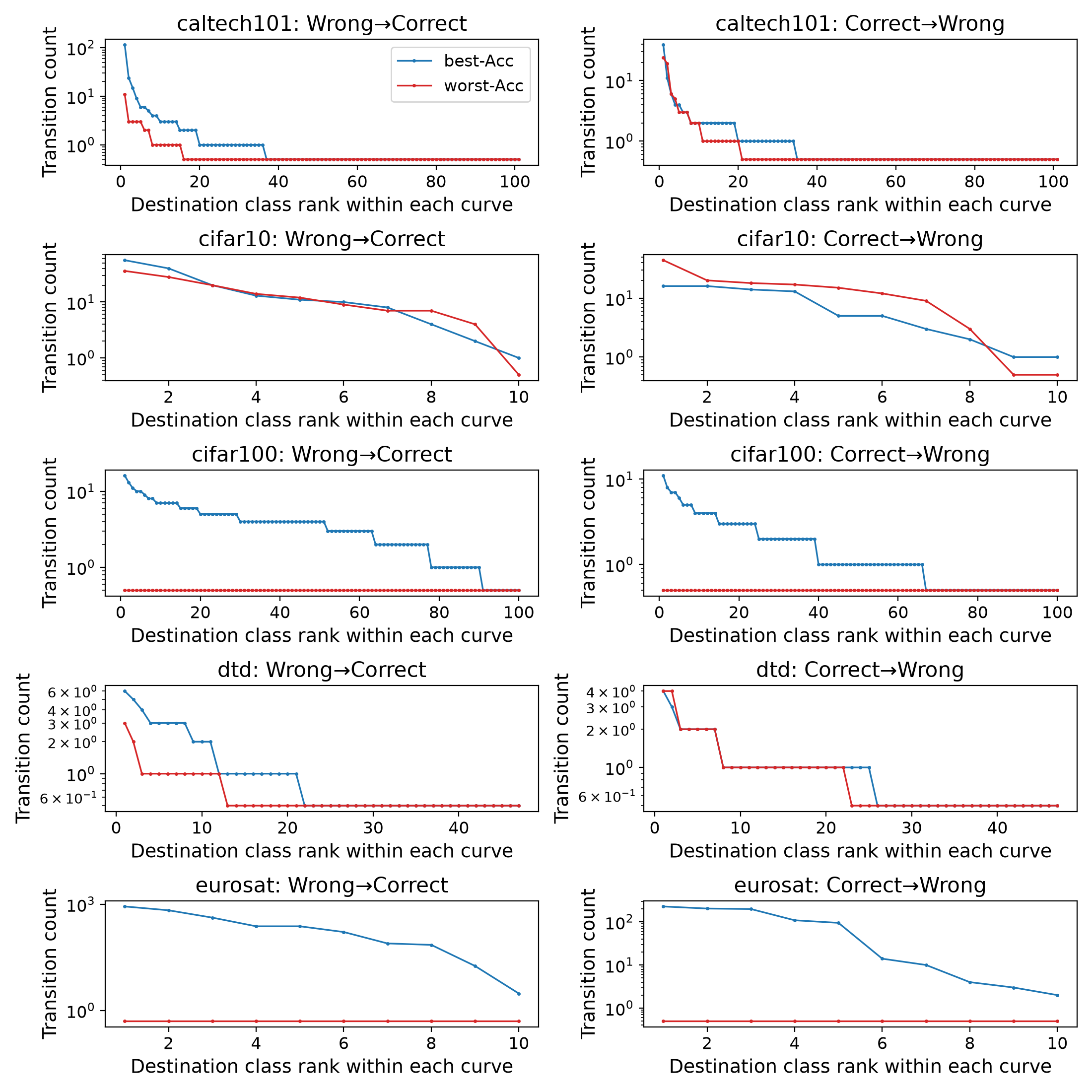}
    \caption{
    Rank curves of destination concentration for prediction transitions under CLIPRefine
    on the first five datasets: Caltech101, CIFAR-10, CIFAR-100, DTD, and EuroSAT.
    Destination classes are independently sorted by transition count within each curve.
    For visualization on the log-scale y-axis, destination classes with zero transitions are shown at 0.5.
    }
    \label{fig:app_prediction_destination_01_cliprefine}
\end{figure*}

\begin{figure*}[t]
    \centering
    \includegraphics[keepaspectratio, scale=0.49]{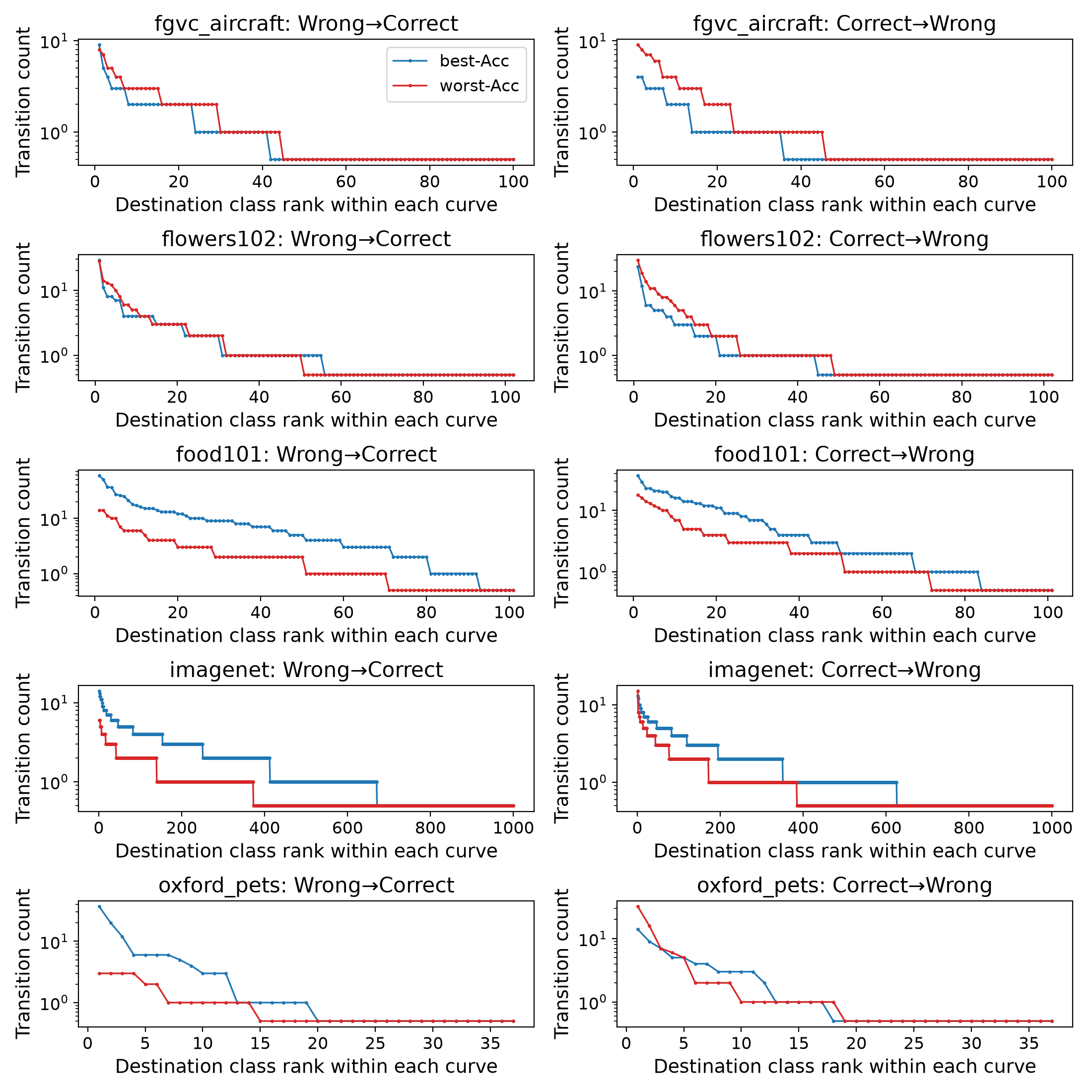}
    \caption{
    Rank curves of destination concentration for prediction transitions under CLIPRefine
    on the remaining five datasets: FGVC-Aircraft, Flowers102, Food-101, ImageNet-1K, and Oxford-IIIT Pet.
    Destination classes are independently sorted by transition count within each curve.
    For visualization on the log-scale y-axis, destination classes with zero transitions are shown at 0.5.
    }
    \label{fig:app_prediction_destination_02_cliprefine}
\end{figure*}

Figures~\ref{fig:app_prediction_destination_01_cliprefine} and~\ref{fig:app_prediction_destination_02_cliprefine} show the corresponding destination-rank curves for CLIPRefine~\cite{CLIPRefine}.
They provide the full transition-level results supporting the comparison between Linear correction and CLIPRefine in the main text.

\subsection{Additional Results for Intervening on Gap-Induced Bias}
\label{app:alpha_sweep}

In Section~\ref{subsec:hub_suppression}, we reported the intervention results at the over-correction setting \(\alpha=0.5\) with \(\lambda=1\).
Here, we provide additional results by sweeping both the Linear correction strength \(\alpha\) and the bias-subtraction strength \(\lambda\).

\begin{figure}[t]
    \centering
    \includegraphics[keepaspectratio, scale=0.33]{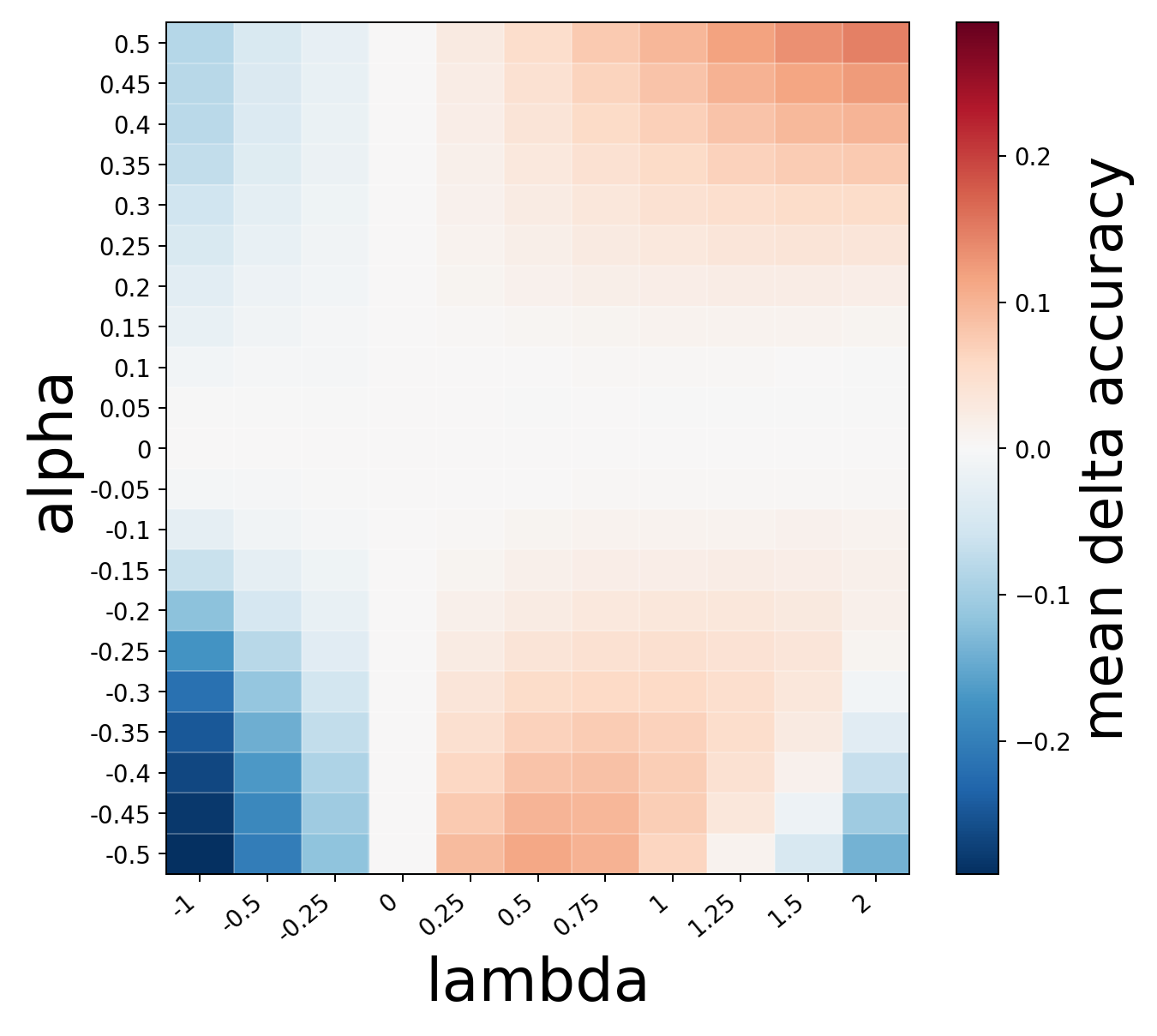}
    \caption{
    Mean change in accuracy after gap-bias subtraction across correction strengths \(\alpha\) and intervention strengths \(\lambda\).
    Values are relative to the Linear-correction baseline at the same \(\alpha\); positive values indicate accuracy improvement.
    }
    \label{fig:alpha_lambda_acc}
\end{figure}

\begin{figure}[t]
    \centering
    \includegraphics[keepaspectratio, scale=0.33]{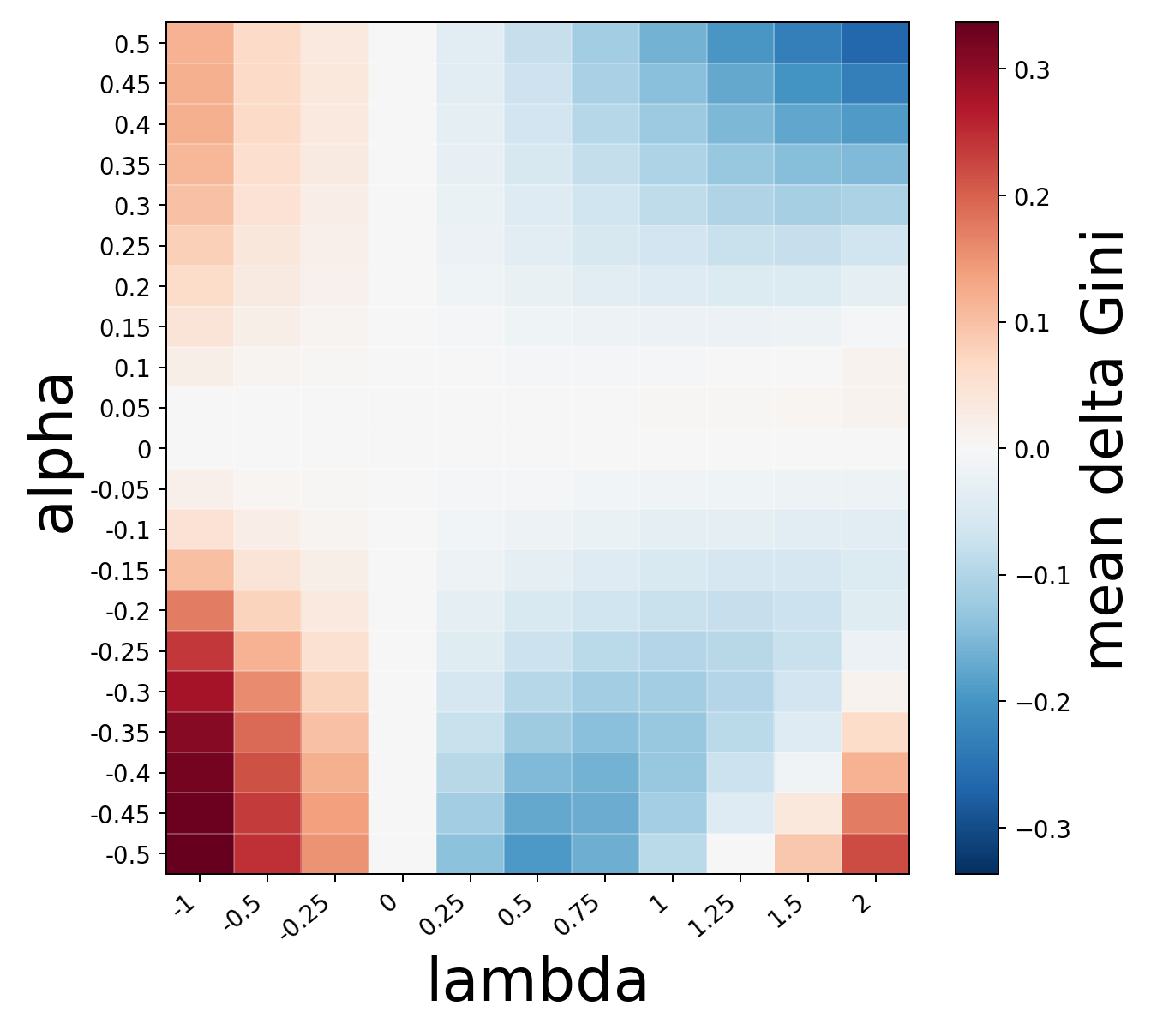}
    \caption{
    Mean change in Predicted-Class Gini after gap-bias subtraction across correction strengths \(\alpha\) and intervention strengths \(\lambda\).
    Values are relative to the Linear-correction baseline at the same \(\alpha\); negative values indicate reduced prediction concentration.
    }
    \label{fig:alpha_lambda_gini}
\end{figure}

Figures~\ref{fig:alpha_lambda_acc} and~\ref{fig:alpha_lambda_gini} show the mean changes in accuracy and Predicted-Class Gini, respectively, after applying the bias-subtraction operation defined in Equation~\ref{eq:bias_subtraction}, which subtracts \(\lambda \alpha b_c\) from each corrected class score.
Each value is computed relative to the corresponding Linear-correction baseline at the same \(\alpha\).
Thus, positive values in Figure~\ref{fig:alpha_lambda_acc} indicate improved accuracy, while negative values in Figure~\ref{fig:alpha_lambda_gini} indicate reduced prediction concentration.

The results show that the effect of bias subtraction depends on both \(\alpha\) and \(\lambda\).
In the positive over-correction regime, positive values of \(\lambda\) generally improve accuracy and reduce Predicted-Class Gini, supporting the interpretation that gap-induced class-wise bias contributes to prediction-level hubness.
In contrast, subtraction with a mismatched direction, such as negative \(\lambda\), can reduce accuracy or increase concentration.
This indicates that the intervention should be understood as a diagnostic test of the proposed mechanism rather than as a general-purpose correction method.

\subsection{Prototype Structure and Uniformity under Correction}
\label{app:uniformity_results}

This section provides the measurements behind the geometric-correlate discussion in Section~\ref{sec:discussion}, using the uniformity definition of Appendix~\ref{app:uniformity_details}.
Prototype--prototype angular structure is essentially preserved by both correction profiles: the correlation between corrected and uncorrected pairwise prototype similarities is 0.994 for Linear correction at \(\alpha=0.5\) and 0.975 for CLIPRefine.
Joint image--text uniformity instead separates the two profiles.
Under Linear correction, it improves only up to \(\alpha \approx 0.10\) and then deteriorates, just past the accuracy peak and coinciding with the rise in Gini (within-dataset Spearman \(-0.92\) with accuracy and \(+0.92\) with Gini, on all 10 datasets).
CLIPRefine instead shrinks the gap by 0.177 while improving uniformity by 0.054, with Gini essentially unchanged.

\section{Robustness Checks and Ablations}
\label{app:robustness_checks}

This appendix collects the robustness checks and ablations referenced in the main text: prompt configuration (Appendices~\ref{app:generic_config}--\ref{app:llm_descriptions}), modality-wise centering (Appendix~\ref{app:centering}), CSLS scoring (Appendix~\ref{app:csls_details}), and model scale (Appendix~\ref{app:model_scale}).

The experiments in the main text use the dataset-specific prompt templates of Table~\ref{tab:app_prompt_templates}; Appendices~\ref{app:generic_config} and~\ref{app:llm_descriptions} examine whether the conclusions depend on that choice, first by re-running all analyses with the generic template \texttt{a photo of a \{class name\}.} applied uniformly to all 10 datasets, and then by replacing hand-written templates entirely with LLM-generated attribute descriptions.

\subsection{Generic-Template Configuration}
\label{app:generic_config}

\begin{table*}[t]
\centering
\small
\setlength{\tabcolsep}{4pt}
\resizebox{\textwidth}{!}{
\begin{tabular}{llcccccccccccc}
\toprule
Metric & Method
& Cal101 & C10 & C100 & DTD & Euro & Aircraft & Flw102 & Food & IN & Pets
& Avg. & Exp. \\
\midrule
\multirow{2}{*}{\(\rho_G\)} & Linear
& -0.819 & -1.000 & -0.997 & -0.997 & -0.203 & -0.944 & -0.926 & -0.997 & -1.000 & -0.995
& -0.888 & 10/10 \\
& CLIPRefine
& 0.797 & -0.928 & -0.910 & -0.298 & -0.843 & -0.215 & -0.280 & -0.850 & -0.728 & -0.701
& -0.496 & 9/10 \\
\midrule
\multirow{2}{*}{\(\rho_H\)} & Linear
& 0.427 & 1.000 & 0.990 & 0.997 & 0.804 & 0.944 & 0.926 & 0.994 & 0.987 & 0.983
& 0.905 & 10/10 \\
& CLIPRefine
& -0.969 & 0.954 & 0.911 & 0.303 & 0.880 & 0.221 & 0.347 & 0.648 & 0.681 & 0.849
& 0.482 & 9/10 \\
\bottomrule
\end{tabular}
}
\caption{
Within-dataset Spearman correlations under the generic-template configuration, in which \texttt{a photo of a \{class name\}.} is used for all 10 datasets.
\(\rho_G\) and \(\rho_H\) denote correlations of accuracy with Predicted-Class Gini and with normalized prediction entropy, respectively; ``Exp.'' counts datasets with the expected direction.
The correlation structure matches Tables~\ref{tab:within_dataset_gini_corr} and~\ref{tab:within_dataset_entropy_corr}: every entry has the same sign in both configurations, including the Caltech101 anomaly discussed in Section~\ref{subsec:reporting}.
}
\label{tab:generic_config}
\end{table*}

Five of the ten datasets (Caltech101, CIFAR-10, CIFAR-100, ImageNet-1K, and Oxford-IIIT Pet) already use the generic template in Table~\ref{tab:app_prompt_templates}, so this check exercises the remaining five (DTD, EuroSAT, FGVC-Aircraft, Flowers102, and Food-101).
Table~\ref{tab:generic_config} reports the within-dataset Spearman correlations of accuracy with Predicted-Class Gini and with normalized prediction entropy under the generic configuration, computed in an independent run (including re-extraction of all embeddings).
Every correlation has the same sign as its counterpart in the main text: Linear correction is in the expected direction on 10 of 10 datasets for both metrics (mean \(\rho_G=-0.888\), against \(-0.922\) with dataset-specific templates), and CLIPRefine on 9 of 10 (mean \(\rho_G=-0.496\), against \(-0.485\)), with Caltech101 remaining the single positive case for the reasons given in Section~\ref{subsec:reporting}.
The pooled Pearson correlations corresponding to Figure~\ref{fig:acc_gini} are \(r=-0.791\) for Linear correction and \(r=-0.739\) for CLIPRefine under the generic configuration, against \(r=-0.778\) and \(r=-0.710\) in the main text.
The transition-level pattern of Table~\ref{tab:transition_destination_concentration} is likewise preserved (for example, 32{,}486 correct-to-wrong transitions in the Linear-worst setting with a top-five destination fraction of 0.660, against 33{,}804 and 0.672 in the main configuration).
The frequency-normalized concentration test of Section~\ref{subsec:reporting}---computing the Gini coefficient over \(r_c=(m_c/N)/(n_c/N)\), i.e.\ each class's predicted-class count divided by its ground-truth frequency, instead of over \(m_c\)---also behaves identically: the Caltech101 correlation reverses from \(+0.797\) to \(-0.981\) and Flowers102, the second most imbalanced dataset, strengthens from \(-0.280\) to \(-0.795\), while the exactly balanced datasets are unchanged by construction and the near-balanced ones keep their negative sign.
We conclude that no result in this paper depends on the prompt-template choice.
The one sign difference we observed anywhere is the EuroSAT accuracy--uniformity correlation noted in Section~\ref{sec:discussion}, which is positive under the generic configuration and negative under the dataset-specific one; EuroSAT is also the dataset with the weakest correlations throughout, and we flag it rather than explain it.

\subsection{LLM-Generated Attribute Descriptions}
\label{app:llm_descriptions}

A remaining question is whether prediction-level hubness is an artifact of short hand-written templates, and whether richer, attribute-level descriptions of the kind used by \citet{saha2024improved} would avoid it.
The mechanism of Section~\ref{sec:mechanism} predicts otherwise: the derivation of \(b_c=-\langle g,t_c\rangle\) makes no reference to the prompt text, which changes only where the prototypes \(t_c\) and the gap vector \(g\) lie, so richer descriptions may change \emph{which} classes become hubs but should not prevent hub formation.
We test this prediction directly as a targeted check on a single dataset, CIFAR-100; we do not claim generality across datasets.

We generate five attribute descriptions per class with Qwen3-4B-Instruct-2507 (greedy decoding), using the prompt format of \citet{saha2024improved}: \emph{``What characteristics can be used to differentiate a \{class\} from other objects based on just a photo? ... Texts should be of the form `a photo of a \{class\} with $\langle$characteristic$\rangle$.'\,''}
We evaluate the descriptions with CLIP ViT-B/32 in two ways, both test-time only with no fine-tuning: scoring against the normalized mean of the five description embeddings per class (mean prototype), and the evaluation rule of \citet{saha2024improved} (their Eq.~3), which averages per-description softmax probabilities within each class.
As a pipeline check, our single-template baseline reproduces the CIFAR-100 row of Table~\ref{tab:linear_bias_hub_corr} to within 0.001 (\(\rho_m=0.870\), \(\rho_{\Delta m}=0.862\), \(\rho_e=0.832\) against 0.869/0.863/0.833).

\begin{table}[t]
\centering
\small
\setlength{\tabcolsep}{4pt}
\begin{tabular}{lcccc}
\toprule
& \multicolumn{2}{c}{Base (\(\alpha=0\))} & \multicolumn{2}{c}{\(\alpha=0.5\)} \\
\cmidrule(lr){2-3}\cmidrule(lr){4-5}
Prompting & Acc. & Gini & Acc. & Gini \\
\midrule
Single template & 62.3 & 0.27 & 48.3 & 0.56 \\
Descriptions (mean proto.) & 63.5 & 0.25 & 46.3 & 0.60 \\
Descriptions (their Eq.~3 rule) & 61.1 & 0.27 & 47.9 & 0.58 \\
\bottomrule
\end{tabular}
\caption{
Linear correction on CIFAR-100 with LLM-generated attribute descriptions.
Over-correction degrades accuracy and roughly doubles Predicted-Class Gini under all three prompting schemes, including the evaluation rule of \citet{saha2024improved}.
}
\label{tab:llm_descriptions}
\end{table}

Table~\ref{tab:llm_descriptions} shows the result: under Linear over-correction, accuracy drops and Predicted-Class Gini roughly doubles in all three schemes, so richer prompts do not remove the failure mode.
Moreover, \(b_c\) remains predictive of hub formation: at \(\alpha=0.5\) the Spearman correlations with predicted-class count, its increase, and correct-to-wrong destinations are 0.833/0.837/0.824 for the mean-prototype scheme (computed with \(t_c\) taken as the mean description embedding) and 0.757/0.757/0.767 under their Eq.~3 evaluation rule.
Consistent with the report of \citet{saha2024improved} that test-time descriptions alone yield modest gains on coarse-grained datasets, base accuracy moves only slightly (62.3\% to 63.5\% for the mean prototype); their fine-tuning setting is outside the scope of this check.

\subsection{Centering Ablation}
\label{app:centering}

\begin{table}[t]
\centering
\small
\setlength{\tabcolsep}{3pt}
\begin{tabular}{lrrrr}
\toprule
& \multicolumn{2}{c}{Acc. range (pt)} & \multicolumn{2}{c}{Gini range} \\
\cmidrule(lr){2-3}\cmidrule(lr){4-5}
Dataset & Linear & +Center & Linear & +Center \\
\midrule
Caltech101    & 28.6 & 0.7 & 0.18 & 0.006 \\
CIFAR-10      &  9.3 & 0.1 & 0.17 & 0.001 \\
CIFAR-100     & 18.2 & 0.1 & 0.34 & 0.009 \\
DTD           & 16.0 & 0.3 & 0.36 & 0.003 \\
EuroSAT       & 13.3 & 0.2 & 0.14 & 0.001 \\
FGVC-Aircraft &  5.2 & 0.2 & 0.23 & 0.005 \\
Flowers102    & 25.3 & 0.1 & 0.41 & 0.002 \\
Food-101      & 19.4 & 0.1 & 0.30 & 0.003 \\
ImageNet-1K   & 27.9 & 0.4 & 0.53 & 0.008 \\
Oxford Pets   & 21.6 & 0.3 & 0.30 & 0.002 \\
\midrule
Mean          & 18.5 & 0.2 & 0.30 & 0.004 \\
\bottomrule
\end{tabular}
\caption{
Range of accuracy (percentage points) and Predicted-Class Gini over the sweep \(\alpha \in [-0.5, 0.5]\), for Linear correction with and without modality-wise centering.
Centering removes the \(\alpha\) dependence of both quantities on every dataset.
}
\label{tab:centering}
\end{table}

This section provides the full results for the modality-wise centering analysis summarized in Section~\ref{subsec:mitigation}.
After Linear correction with strength \(\alpha\), we apply, in order: the shift by \(-\alpha g\) (image side; \(+\alpha g\) on the text side), per-sample L2 normalization, subtraction of the modality mean, and renormalization.
Because centering removes each modality's mean, it removes the residual gap direction and with it the class-wise bias \(b_c\); the prediction of Section~\ref{sec:mechanism} is therefore that accuracy and concentration become (approximately) independent of \(\alpha\).

Table~\ref{tab:centering} confirms this on every dataset.
The per-dataset accuracy range across the full sweep drops from 18.5 points on average (maximum 28.6) to at most 0.7 points, and the Predicted-Class Gini range from 0.30 on average (maximum 0.53) to at most 0.009.
The residual \(\alpha\) dependence is explained by the per-sample normalization between the shift and the centering, which makes the two operations non-commuting; the correction is exactly cancelled only without that intermediate normalization.
Under the generic-template configuration of Appendix~\ref{app:generic_config}, the same pattern holds, with accuracy ranges below 0.8 points and Gini ranges below 0.011 on every dataset.

Centering is a diagnostic, not a practical improvement: its effect on the accuracy of the uncorrected model is dataset-dependent, from \(-4.2\) points (Flowers102) to \(+4.1\) points (EuroSAT) under the dataset-specific templates, and up to \(+15.3\) points on EuroSAT under the generic template.
What the ablation establishes is that the entire \(\alpha\)-dependent failure mode, both the accuracy degradation and the prediction concentration, disappears when the gap direction is removed, which is the embedding-space counterpart of the score-space intervention in Section~\ref{subsec:hub_suppression}.


\subsection{CSLS Scoring}
\label{app:csls_details}

\begin{table*}[t]
\centering
\small
\setlength{\tabcolsep}{3pt}
\resizebox{\textwidth}{!}{
\begin{tabular}{lcccccccccccc}
\toprule
Metric
& Cal101 & C10 & C100 & DTD & Euro & Aircraft & Flw102 & Food & IN & Pets
& Avg. & Exp. \\
\midrule
\(\rho_G\)
& -0.918 & -0.978 & -0.995 & -0.956 & -0.929 & -0.822 & -0.982 & -0.982 & -0.992 & -0.842
& -0.940 & 10/10 \\
\(\rho_H\)
& 0.567 & 0.998 & 0.962 & 0.948 & 0.875 & 0.834 & 0.982 & 0.944 & 0.992 & 0.729
& 0.883 & 10/10 \\
\bottomrule
\end{tabular}
}
\caption{
Within-dataset Spearman correlations for Linear-CSLS.
\(\rho_G\) and \(\rho_H\) measure correlations with Predicted-Class Gini and normalized entropy, respectively.
``Exp.'' indicates the number of datasets with the expected correlation direction.
}
\label{tab:within_dataset_corr_csls}
\end{table*}

\begin{figure}[t]
    \centering
    \includegraphics[keepaspectratio, scale=0.33]{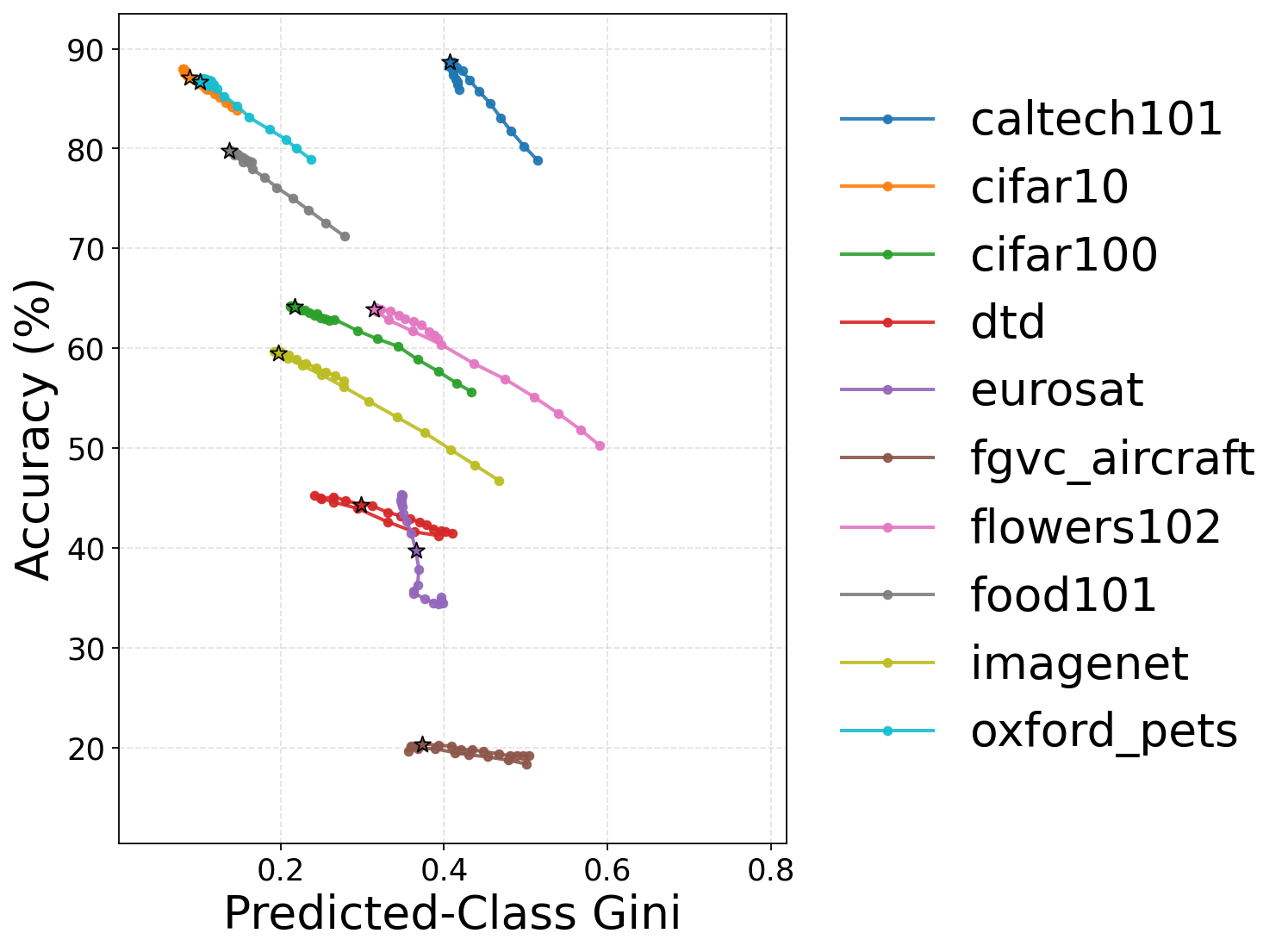}
    \caption{
    Predicted-Class Gini vs. accuracy under Linear correction with CSLS.
    Higher Predicted-Class Gini generally corresponds to lower accuracy, indicating that prediction concentration remains associated with accuracy degradation under CSLS.
    Stars indicate the base model.
    }
    \label{fig:gini_csls}
\end{figure}
\begin{figure}[t]
    \centering
    \includegraphics[keepaspectratio, scale=0.33]{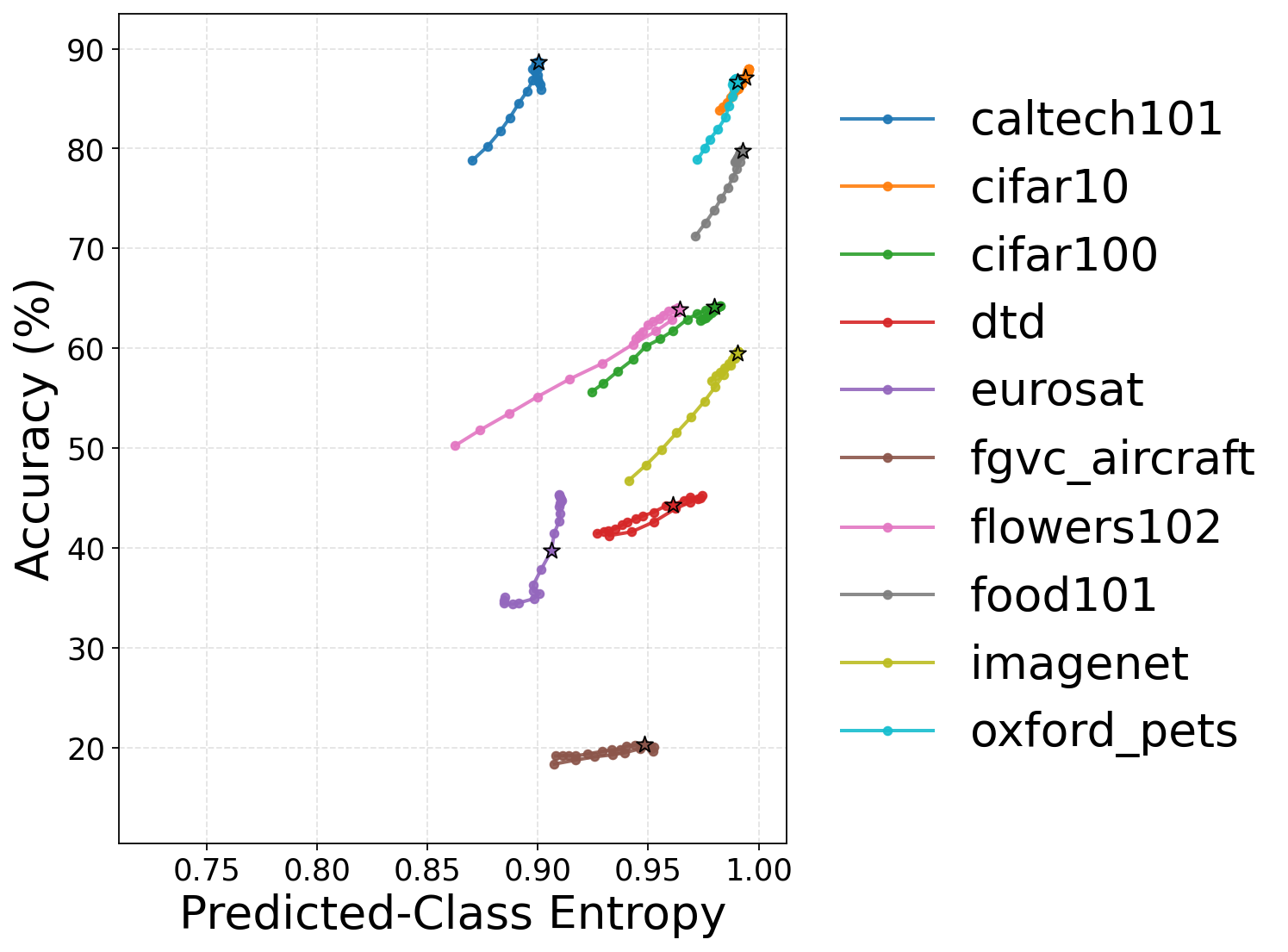}
    \caption{
    Normalized prediction entropy vs. accuracy under Linear correction with CSLS.
    Higher normalized prediction entropy generally corresponds to higher accuracy, showing the complementary trend to Predicted-Class Gini.
    Stars indicate the base model.
    }
    \label{fig:entropy_csls}
\end{figure}

Section~\ref{subsec:mitigation} evaluates Linear correction under Cross-domain Similarity Local Scaling (CSLS), a local-scaling similarity originally introduced for cross-lingual embedding retrieval~\cite{CSLS} and commonly used in settings where nearest-neighbor hubness is a concern~\citep{hubs-in-space,zero-shot-hub,balance-act,NeighborRetr}.
Beyond its role as a partial mitigation, CSLS also addresses the concern that the concentration--accuracy relationship might be tied to raw cosine similarity.
This section provides the scoring definition and the full per-dataset results.

\paragraph{CSLS scoring.}
Let \(x_i \in \mathbb{R}^{d}\) be the image embedding of sample \(i\), and let \(t_c \in \mathbb{R}^{d}\) be the text prototype of class \(c\).
All embeddings are L2-normalized.
The raw cosine similarity is
\[
s(x_i,t_c)=x_i^\top t_c .
\]
For each image embedding \(x_i\), we define its average similarity to the top-\(k\) nearest text prototypes as
\[
r_X(x_i)
=
\frac{1}{k}
\sum_{t\in \mathcal{N}_T^k(x_i)}
s(x_i,t),
\]
where \(\mathcal{N}_T^k(x_i)\) denotes the set of top-\(k\) text prototypes nearest to \(x_i\).
Similarly, for each text prototype \(t_c\), we define
\[
r_T(t_c)
=
\frac{1}{k}
\sum_{x\in \mathcal{N}_X^k(t_c)}
s(x,t_c),
\]
where \(\mathcal{N}_X^k(t_c)\) denotes the set of top-\(k\) image embeddings nearest to \(t_c\).
We set \(k=10\) in all CSLS experiments.

The CSLS score is
\[
\operatorname{CSLS}(x_i,t_c)
=
2s(x_i,t_c)
-
r_X(x_i)
-
r_T(t_c).
\]
The predicted class is obtained by
\[
\hat{y}_i
=
\arg\max_c
\operatorname{CSLS}(x_i,t_c).
\]

\paragraph{Linear correction with CSLS.}
For Linear correction with CSLS, we apply the same embedding-shift formulation as in the main Linear correction experiments~\cite{modality-gap}.
We use the same gap vector:
\[
g
=
\frac{1}{N}\sum_{i=1}^{N}x_i
-
\frac{1}{N}\sum_{j=1}^{N}t_j .
\]
For correction strength \(\alpha\), we shift image and text embeddings in opposite directions and renormalize them:
\[
\tilde{x}_i
=
\frac{x_i-\alpha g}{\|x_i-\alpha g\|_2},
\qquad
\tilde{t}_c
=
\frac{t_c+\alpha g}{\|t_c+\alpha g\|_2}.
\]
We then compute CSLS using \(\tilde{x}_i\) and \(\tilde{t}_c\) in place of \(x_i\) and \(t_c\).

\paragraph{Results.}
Compared with raw cosine scoring, CSLS roughly halves the over-correction penalty: at \(\alpha=0.5\), the mean accuracy change from the base model is \(-7.4\) points under CSLS versus \(-15.1\) points under cosine scoring.
Figures~\ref{fig:gini_csls} and~\ref{fig:entropy_csls} further show that prediction concentration remains associated with accuracy under CSLS scoring.
Higher Predicted-Class Gini generally corresponds to lower accuracy, while normalized prediction entropy shows the complementary trend.
Thus, CSLS reduces but does not fully remove the prediction-level concentration induced by excessive gap correction.
These results support the interpretation that gap correction can create a hubness-related failure mode, rather than merely changing the average modality gap or depending on a particular raw similarity function.

\begin{figure}[t]
    \centering
    \includegraphics[keepaspectratio, scale=0.31]{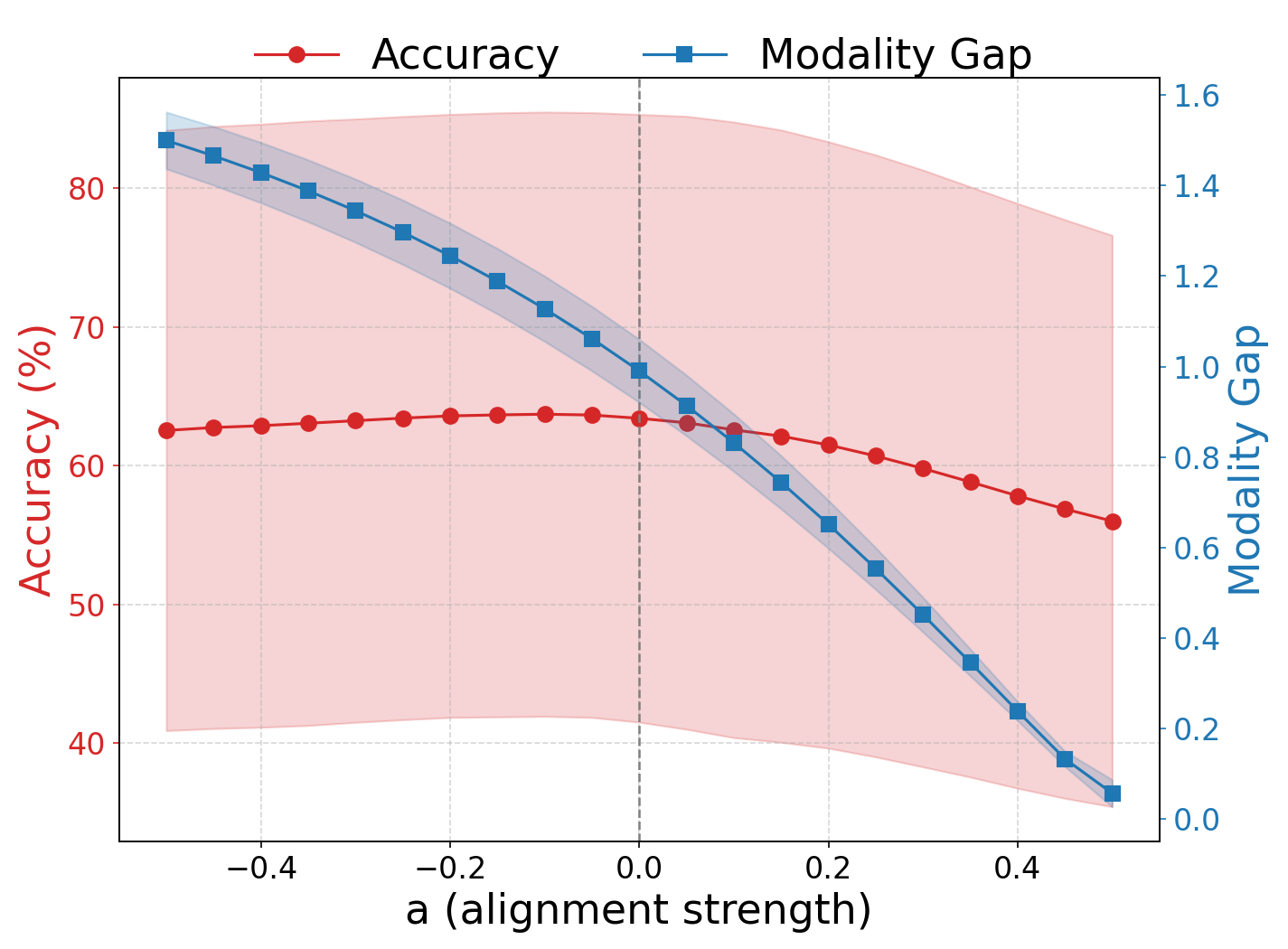}
    \caption{
    Accuracy and modality gap under Linear correction with CSLS scoring.
    The modality gap decreases monotonically with correction strength, whereas accuracy improves only up to a point and then deteriorates.
    }
    \label{fig:linear_csls}
\end{figure}

Figure~\ref{fig:linear_csls} shows accuracy and modality gap under Linear correction with CSLS scoring.
CSLS partially mitigates the degradation caused by strong Linear correction, but does not eliminate it.
As the correction strength increases, the modality gap still decreases monotonically, while accuracy improves only up to a point and then deteriorates.
This suggests that part of the degradation under gap correction is related to hubness in the similarity-based prediction rule, rather than being an artifact of raw cosine similarity alone.

Table~\ref{tab:within_dataset_corr_csls} reports within-dataset Spearman correlations between accuracy and prediction concentration.
Linear-CSLS shows the expected direction in all 10 datasets, with a negative correlation for Predicted-Class Gini and a positive correlation for normalized entropy.
This indicates that the concentration--accuracy relationship is not merely a pooled cross-dataset artifact and remains visible under CSLS scoring.

\subsection{Effect of Model Scale}
\label{app:model_scale}

\begin{table*}[t]
\centering
\small
\setlength{\tabcolsep}{4pt}
\resizebox{\textwidth}{!}{
\begin{tabular}{lcccccccccccc}
\toprule
Backbone
& Cal101 & C10 & C100 & DTD & Euro & Aircraft & Flw102 & Food & IN & Pets
& Avg. & Exp. \\
\midrule
ViT-B/32
& -0.819 & -0.998 & -0.997 & -0.996 & -0.725 & -0.766 & -0.934 & -0.995 & -1.000 & -0.993
& -0.922 & 10/10 \\
ViT-B/16
& -0.907 & -0.900 & -0.987 & -0.963 & 0.903 & -0.957 & -0.977 & -0.996 & -0.995 & -0.986
& -0.776 & 9/10 \\
ViT-L/14
& -0.806 & -0.922 & -0.997 & -0.994 & -0.930 & -0.981 & -0.587 & -0.992 & -1.000 & -0.994
& -0.920 & 10/10 \\
\bottomrule
\end{tabular}
}
\caption{
Within-dataset Spearman correlations between accuracy and Predicted-Class Gini for the Linear-correction sweep across CLIP backbones.
``Exp.'' counts datasets with the expected negative direction.
The ViT-B/32 row repeats Table~\ref{tab:within_dataset_gini_corr} for comparison.
}
\label{tab:model_scale}
\end{table*}

One might hope that larger CLIP models, whose modality gap may be smaller or more uniform, do not exhibit prediction-level hubness.
To test this, we repeat the full Linear-correction sweep (21 correction strengths, all 10 datasets, dataset-specific templates) on ViT-B/16 and ViT-L/14, with the gap vector estimated per backbone as in Appendix~\ref{app:linear_correction_details}.

Table~\ref{tab:model_scale} shows that the phenomenon does not diminish with scale.
On ViT-L/14 the correlation between accuracy and Predicted-Class Gini is negative on all 10 datasets with a mean of \(-0.920\), matching ViT-B/32 (\(-0.922\)), and the pooled Pearson correlation is in fact strongest on the largest backbone (\(-0.815\), against \(-0.778\) for B/32 and \(-0.791\) for B/16).
Over-correction remains costly: at \(\alpha=0.5\), ViT-L/14 loses 10.5 points of accuracy on average, and 22.0 points on ImageNet-1K, where the correlation is \(-1.000\).
This is consistent with the mechanism: the derivation of \(b_c=-\langle g,t_c\rangle\) is backbone-independent, so a smaller or differently oriented gap changes the size of the induced bias but not the structure of the failure mode.
The single exception is EuroSAT on ViT-B/16 (\(+0.903\)), where over-correction happens to improve accuracy; EuroSAT has only 10 classes, so its Gini values span a narrow range and the rank correlation is unstable, and the same dataset is the weakest for ViT-B/32 while returning to \(-0.930\) on ViT-L/14.

\end{document}